%% file: main.tex
\documentclass[lettersize,journal]{IEEEtran}
\usepackage{amsmath,amsfonts}
\usepackage{algorithmic}
\usepackage{array}
\usepackage[caption=false,font=normalsize,labelfont=sf,textfont=sf]{subfig}
\usepackage{textcomp}
\usepackage{stfloats}
\usepackage{url}
\usepackage{verbatim}
\usepackage{graphicx}
\usepackage{amssymb}
\usepackage[switch]{lineno}
\usepackage{cite}

\def\BibTeX{{\rm B\kern-.05em{\sc i\kern-.025em b}\kern-.08em
    T\kern-.1667em\lower.7ex\hbox{E}\kern-.125emX}}
\usepackage{balance}
\usepackage{multirow}

\usepackage[colorlinks,
            linkcolor=blue,
            anchorcolor=blue,
            citecolor=blue]{hyperref}

\begin{document}

\title{RGB-Sonar Tracking Benchmark and Spatial Cross-Attention Transformer Tracker}

\author{Yunfeng Li, Bo Wang*, Jiuran Sun, Xueyi Wu, Ye Li.
\thanks{* Corresponding author. Email: cv\_heu@163.com.} }

\maketitle
\input{sections/0_abstract}
\input{sections/0_keywords}
\input{sections/1_introduction}
\input{sections/2_related_work}
\input{sections/3_benchmark}
\input{sections/4_method}
\input{sections/5_experiments}
\input{sections/6_conclusion}
\input{sections/acknowledgement}

\bibliography{references}
\bibliographystyle{IEEEtran}

\end{document}

%% file: sections/0_abstract.tex
\begin{abstract}
Vision camera and sonar are naturally complementary in the underwater environment. Combining the information from two modalities will promote better observation of underwater targets. However, this problem has not received sufficient attention in previous research. Therefore, this paper introduces a new challenging RGB-Sonar (RGB-S) tracking task and investigates how to achieve efficient tracking of an underwater target through the interaction of RGB and sonar modalities. Specifically, we first propose an RGBS50 benchmark dataset containing 50 sequences and more than 87000 high-quality annotated bounding boxes. Experimental results show that the RGBS50 benchmark poses a challenge to currently popular SOT trackers. Second, we propose an RGB-S tracker called SCANet, which includes a spatial cross-attention module (SCAM) consisting of a novel spatial cross-attention layer and two independent global integration modules. The spatial cross-attention is used to overcome the problem of spatial misalignment of between RGB and sonar images. Third, we propose a SOT data-based RGB-S simulation training method (SRST) to overcome the lack of RGB-S training datasets. It converts RGB images into sonar-like saliency images to construct pseudo-data pairs, enabling the model to learn the semantic structure of RGB-S-like data. Comprehensive experiments show that the proposed spatial cross-attention effectively achieves the interaction between RGB and sonar modalities and SCANet achieves state-of-the-art performance on the proposed benchmark. The code is available at \url{https://github.com/LiYunfengLYF/RGBS50}.

\end{abstract}  

%% file: sections/0_keywords.tex

%% file: sections/1_introduction.tex
\section{Introduction}

`Underwater Camera and Forward-Looking Sonar are important sensors for underwater observation. They have a wide range of potential applications in underwater monitoring \cite{underwater_monitoring}, marine aquaculture \cite{marine_aquaculture}, underwater vehicle perception \cite{underwater_vehicle_perception}.

Visual RGB images are rich in semantic information, but light scattering by water limits the camera's field of view and leads to underwater image distortion. Despite significant advances in visual single object tracking (SOT), it remains a challenge in underwater image distortion and color degradation scenarios. Sonar images have a longer perceptual range and better robustness than RGB images, but due to the imaging principle, there is little semantic information about the target in a sonar image, which limits the ability of the tracker to discriminate based on appearance features. Overall, single modality information is not sufficient to achieve stable and accurate tracking of underwater targets.

Considering that the perceptual range and semantic structure between the visual (RGB) modality and the sonar modality are complementary. This paper aims to explore how to achieve accurate and robust tracking of underwater targets through cross-modal interaction between RGB and sonar information, named RGB-Sonar (RGB-S) tracking.

Current visual underwater object tracking (UOT) has some tracking benchmarks like UOT100 \cite{uot100}, UTB180 \cite{utb180}, etc. They are a great boost to the development of the underwater tracking community. For sonar tracking, which is limited by the high cost of sonar equipment and experiments, to the best of our knowledge, there are currently no benchmark datasets suitable for sonar tracking. Overall, while there has been some progress in underwater tracking, there is currently no paired RGB-S tracking benchmark that provides a reference for the development of underwater multimodal trackers.

Other multimodal visual tasks such as RGB-Thermal (RGB-T) tracking, RGB-Depth (RGB-D) tracking, etc., are based on the assumption of temporal alignment and spatial alignment of the two modal images. Different with these multimodal tasks, the assumption of spatial alignment is not valid in RGB-S tracking due to the characteristics of sonar images, i.e., pixel points at the same position in paired RGB and sonar images do not represent similar target semantics, as shown in Figure \ref{fig: rgb_tdes}. This distinguishes the RGB-S tracker from other multimodal trackers.

\input{figures/fig_rgb_tdes}

Based on the above analysis, the core issues of this paper are described as the following two points:
\begin{itemize}
\item[$\bullet$] How to create a comprehensive RGB-S tracking benchmark dataset to promote the development of underwater multimodal trackers.
\item[$\bullet$] How to develop an RGB-S tracker in the case of spatial misalignment between RGB and sonar images and lack of RGB-S paired training data.

\end{itemize}

To answer the above questions, in this paper, we propose the first RGB-S tracking benchmark dataset, named RGBS50. The RGBS50 dataset contains 50 time-aligned underwater visual and sonar videos and more than 87000+ high-quality annotations. These paired videos are collected in a deep water pool. We evaluate 25 popular SOT trackers on the proposed benchmark to promote the development of underwater RGB-S tracking.

In addition, we propose an RGB-S tracker SCANet. It achieves efficient interaction between spatially non-aligned RGB and sonar modalities by using a novel spatial cross-attention module (SCAM), which includes a spatial cross-attention block and two independent global integration modules (GIM). In each modality branch, the SCAM first computes the spatial attention map by its own Q-matrix and the K-matrix of another modality, and then implements the filtering of the associated information by the ReLU function, and finally multiplies the spatial attention maps by the V-matrix to obtain the cross-modal feature. It then uses independent GIM to globally integrate each modality feature.

To train the proposed SCANet, we propose a SOT data-based RGB-S simulation training method (SRST). It collects two templates and two search areas in the same sequence at the same time, and then converts one of the paired template and search area images into grayscale saliency images as input to the sonar branch, and inputs the other RGB images to the RGB branch. In addition, it introduces synthetic aperture radar (SAR) detection images to facilitate the model's learning of the low-level semantic structure of the sonar-like images.

The main contributions of this paper are summarized as follow:

\begin{itemize}
\item[$\bullet$] We create the first underwater RGB-Sonar (RGB-S) tracking benchmark, RGBS50, including 50 underwater video sequences and high-quality annotations. We also evaluate 25 popular SOT trackers as baselines and report their performance. This benchmark will promote the development of underwater RGB-S trackers

\item[$\bullet$] We propose a novel spatial cross-attention module named SCAM. It models the feature associations when the target is spatially misaligned by computing the spatial attention map, filters out the background correlations between two modalities using the ReLU function, and then globally integrates the features of the two modalities separately.

\item[$\bullet$] We propose a novel SOT data-based RGB-S simulation training method called SRST. It simulates the spatial misalignment of RGB-S tracking by collecting two different images of the same sequence in a single sampling, and then simulates the sonar image by transforming the second image into a gray-scale saliency image. It also introduces SAR images similar to sonar images to simulate the learning of the low-level semantics of sonar targets.

\item[$\bullet$] We propose an RGB-S tracker named SCANet. It inserts the SCAM modules into different layers of ViT to achieve multi-level feature interaction between RGB and sonar modalities. It completes the training with the RGB-based fine-tuning method. Comprehensive experiments show that our method achieves state-of-the-art performance.

\end{itemize}

%% file: figures/fig_rgb_tdes.tex
\begin{figure}
	\centering
        \includegraphics[width=8.8cm]{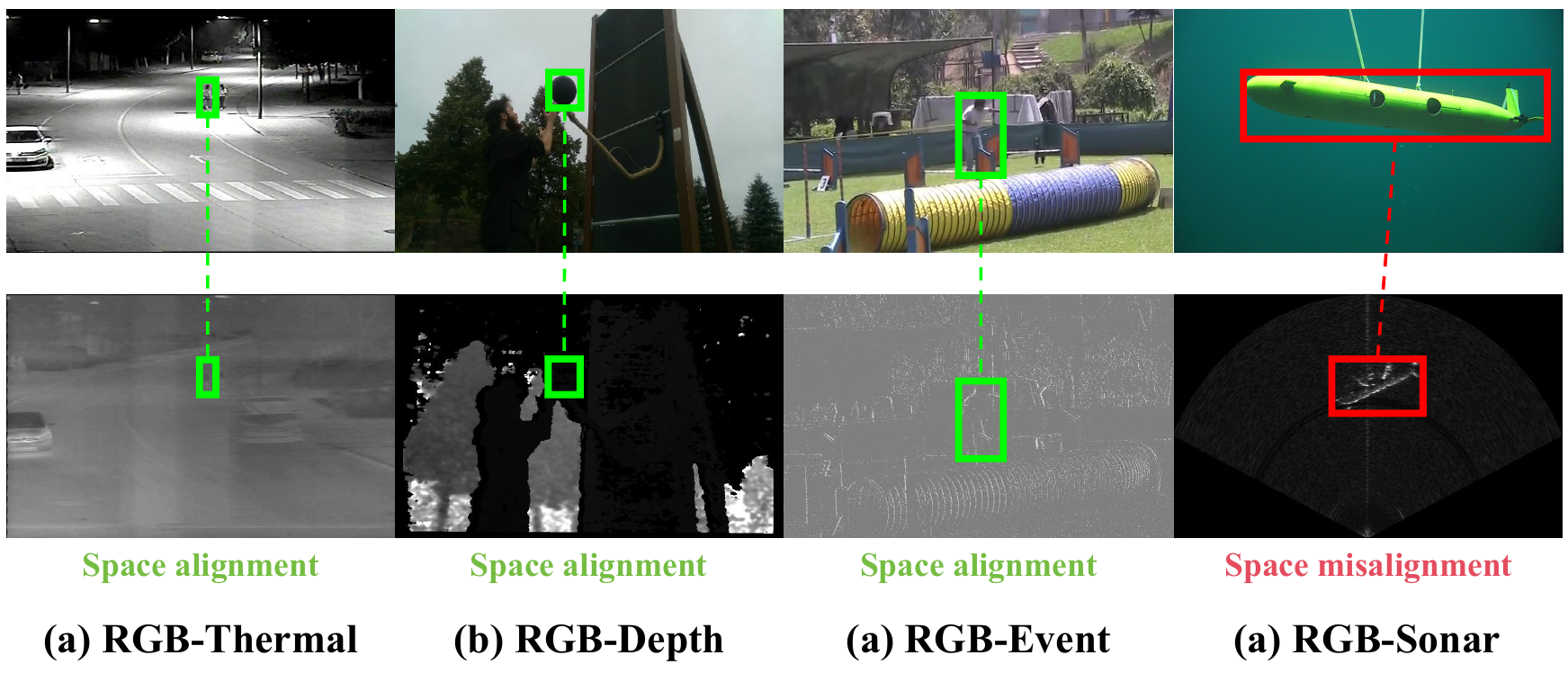}
	\caption{Comparison of the RGB-S tracking task with other multimodal tracking tasks. The RGB images and sonar images have spatial misalignment characteristic. (a) Paired RGB-T images of LasHeR \cite{lasher}, (b) Paired RGB-D images of DepthTrack \cite{depthtrack}, (c) Paired RGB-E images of VOT2019-RGBE \cite{vot2019rgbe}, (d) Paired RGB-S images of our proposed dataset.}
	\label{fig: rgb_tdes}
\end{figure}

%% file: sections/2_related_work.tex
\section{Related Work}

\subsection{Single Object Tracking}
Single Object Tracking (SOT) is a typical task in computer vision and has been widely studied. Some typical correlation filter-based trackers include KCF \cite{kcf}, ECO \cite{eco}, etc. They have been widely applied in various research fields, such as underwater tracking. Siamese-based trackers achieve appearance feature discrimination based on depth feature. Some typical Siamese-based trackers include SiamFC \cite{siamfc}, which establishes the Siamese-based tracker paradigm, SiamFPN \cite{siamrpn}, which establishes the anchor-based prediction head paradigm. In addition, SiamRPN++ \cite{siamrpn++} introduces a deeper backbone network. SiamCAR \cite{siamcar} and SiamBAN \cite{siamban} achieve more accurate anchor-free prediction heads. Transformer-based tracker achieves better global feature modeling. Stark \cite{stark} extracts features from ResNet and models relationships using Transformer blocks. TransT \cite{transt} proposes Transformer-like modules ECA and CFA to model the relationship between template and search area features. OSTrack \cite{ostrack} embeds and concatenates the template and the search area image patches into tokens and feeds them into ViT for joint feature extraction and relationship modeling. In addition, more powerful Transformer trackers such as HIPTrack \cite{hiptrack}, ODTrack \cite{odtrack}, etc. are proposed to further promote the development of SOT.

Overall, these methods achieve high performance tracking in single modality images. They also provide reference and pre-trained models for the design of multimodal trackers.

\subsection{Underwater Tracking}

The underwater RGB image is very different from the sonar image. Therefore, underwater tracking is divided into Underwater Object Tracking (UOT) and Sonar Tracking. 
As a subtask of SOT, UOT aims to address the challenges of underwater image distortion and typical tracking challenges in underwater scenes. UOT100 \cite{uot100} provides a benchmark for evaluating underwater trackers to address the challenge of underwater color distortion, UTB180 \cite{utb180} provides a benchmark to address typical tracking challenges in underwater environments. In addition, some large-scale UOT datasets are also proposed, such as UVOT400 \cite{uvot400} and WebUOT \cite{webuot}. Based on these benchmarks, some underwater trackers are proposed. UStark \cite{ustark} proposes an underwater adaptive enhancement method to reduce the impact of underwater image distortion on tracking performance. UOSTrack \cite{uostrack} proposes UOHT and MBPP methods, respectively, to address class imbalance and similar appearance challenges in UOT. Sonar tracking is similar to UOT. It aims to track a single object moving in the sonar. Due to the lack of publicly available sonar tracking benchmark datasets, there are currently few relevant sonar trackers available.

Compared to these single modality UOT benchmarks, our proposed RGBS50 benchmark provides a reference for the development of multi-mode underwater trackers. Compared to current underwater trackers, our proposed SCANet achieves better tracking performance through cross-modal interaction.

\subsection{Multi-Modal Tracking}

The introduction of Transformer has greatly facilitated the development of multimodal tracking. ViPT \cite{vipt} proposes a multimodal prompt learning paradigm based on the SOT model. TBSI \cite{tbsi} proposes an attention-based template bridging search area feature interaction module that is inserted into different layers of OSTrack \cite{ostrack}. BAT \cite{bat} proposes a linear layer-based bidirectional adapter to achieve cross-modal feature interactions at each layer in ViT. TATrack \cite{tatrack} achieves cross-modal feature interaction by introducing an online template and prompt learning. MPLT \cite{mplt} introduces a mutual prompt learning method between the two modalities by designing a lightweight prompter. CSTNet \cite{cstnet} proposes spatial and channel feature fusion modules for direct cross-modal feature interaction. 

Overall, these methods use a prediction head based on the assumption that the two modal spaces are aligned to predict the same box of the target in both modalities. However, in RGB-S tracking, the above multimodal tracking is not directly applicable to RGB-S tracking because the target's box is different in the two modalities. In addition, compared to these methods, we explore a cross-modal feature relationship modeling method when multimodal features are spatially misaligned.

%% file: sections/3_benchmark.tex
\section{RGB-Sonar Tracking Benchmark}

\subsection{RGBS50 Benchmark}

\input{figures/fig_rgbs50_benchmark}

We collect 50 paired underwater RGB and sonar video sequences. During data collection, we float the targets in the pool and move them around by dragging them. During data processing, we manually annotate all target bounding boxes, and each bounding box annotation is reviewed and corrected by a full-time annotator to ensure accuracy. 
The details of the RGBS50 benchmark are as follows:

\textbf{Hardware Setup}. Our multi-sensor platform is shown in Figure \ref{fig: benckmark}(a). The platform consists of an HD Zoom SeaCam underwater camera and an Oculus MD-Series sonar. The sonar operating mode is high frequency. The sampling frequency of the sensors is uniformly set to 10 frames per second.

\textbf{Annotation}. We align the RGB and sonar images in time and manually annotate the target in each frame of the RGB and sonar sequences. If there is no target in the current frame, the bounding box is annotated as [0,0,0,0]. If the target is occluded, the bounding box is annotated as only the visible part of the target. The total number of annotations on the bounding box is 87404. All inaccurate bounding boxes are checked and corrected by an annotator.

\textbf{Statistics}. We analyze the statistics of our dataset from the following points.

\input{tables/attribute}

\input{figures/fig_rgbs50_examples}

\begin{itemize}
\item[$\bullet$] \textit{Object categories}: Our benchmark contains 7 typical underwater targets, which are ball and polyhedron, connected polyhedron, fake person, frustum, iron ball, octahedron and uuv. The percentage distribution is shown in Figure \ref{fig: benckmark}(b).

\item[$\bullet$] \textit{Attributes}: Our benchmark contains 9 different attributes:  Occlusion (OC), Full Out-of-View (FOV), Similar Appearance (SA), Scale Variation (SV), Sonar Crossover (SC), Deformation (DEF), Visual Low Resolution  (VLR), Low Illumination (LI), Low Sonar Reflection (LSR). Due to the different perceptual ranges of underwater cameras and sonar, it is common in RGB-S tracking for the target to be out of view in one modality and in view in the other. Therefore, we will not discuss the single modality out of view attribute. The detailed definitions of these attributes are shown in Table \ref{table attribute}. The statistics are shown in Figure \ref{fig: benckmark}(c). Overall, our attributes are evenly distributed at the frame and sequence levels.

\item[$\bullet$] \textit{Sequences}: The proposed benchmark contains a total of 50 sequences with a total number of frames of 43702 and a total number of annotations of 87404. Among them, the longest sequence contains 2740 frames, and the shortest sequence contains 251 frames. On average, the sequences have 874 frames. Our proposed RGBS50 dataset is compared with underwater tracking benchmark datasets and multimodal benchmark datasets of the same size, as shown in Table \ref{table benchmark_compare}.

\item[$\bullet$] \textit{Scene Type}: Due to the limitations of the water pool, the scene of RGB sequences is relatively simple. Our dataset includes only normal scene and low light scene in RGB modality. Due to the imaging characteristics of sonar, sonar images do not distinguish scenes. Overall, our dataset reflects the challenge of fusion of non-aligned RGB-S images rather than simply the challenge of object tracking.

\end{itemize}

To the best of our knowledge, there is currently no publicly available benchmark dataset for RGB-S tracking. Our proposed RGB-S50 dataset will facilitate the development of underwater RGB-S multimodal tracking.

\input{tables/benchmarks_compare}

\subsection{Baseline Methods}

In order to facilitate the development of RGB-S tracking, we first evaluated 25 popular SOT trackers on the proposed benchmark. The trackers contains 5 Siamese trackers SiamRPN \cite{siamrpn}, SiamRPN++ \cite{siamrpn++}, SiamBAN \cite{siamban}, SiamCAR \cite{siamcar}, SiamBAN-ACM \cite{siambanacm}, 7 DCF trackers DiMP18 \cite{dimp}, DiMP50 \cite{dimp}, PrDiMP18 \cite{prdimp}, PrDiMP50 \cite{prdimp}, KeepTrack \cite{keeptrack}, TrDiMP \cite{trdimp}, ToMP50 \cite{tomp} and 13 transformer trackers Stark-S50 \cite{stark}, Stark-ST50 \cite{stark}, Stark-ST101 \cite{stark}, TransT \cite{transt}, SLT-TransT \cite{slttranst}, AiATrack \cite{aiatrack}, CSWinTT \cite{cswintt}, OSTrack256 \cite{ostrack}, OSTrack384 \cite{ostrack}, SeqTrack \cite{seqtrack}, ARTRackSeq \cite{artrackv2}, ODTrack \cite{odtrack}, HIPTrack \cite{hiptrack}. They are evaluated independently in RGB and sonar sequences. Results from two modality sequences are analyzed and reported separately.

Since current RGB-T, RGB-D, RGB-E trackers are based on a multimodal image space alignment assumption, they use a prediction head to predict the state of the target. This prevents these trackers from being directly applied to RGB-S tracking.

\subsection{Evaluation Metrics}
We follow the One-Pass Evaluation (OPE) protocol to evaluate the baseline trackers. Then, we use three evaluation metrics - Precision (PR), Norm Precision (NPR), and Success Rate (SR) - that are widely used in the current tracking community to evaluate tracking performance. 
\begin{itemize}
\item[$\bullet$] \textbf{Precision Rate (PR)}. We calculate the percentage of frames where the distance between the predicted position and the ground truth is within a certain threshold range, then obtain the PR score by setting the threshold to 20.

\item[$\bullet$] \textbf{Normalized precision Rate (NPR)}. Following the setting of \cite{lasher}, we introduce NPR to avoid the impact of the image resolution and the bounding box size on the PR.

\item[$\bullet$] \textbf{Success Rate (SR)}. We obtain the ratio of successful frames by calculating the overlap rate between the ground true and predicted boxes that is greater than a certain threshold. Then we obtain the SR score through the area under the curve.
\end{itemize}

Since the RGB and sonar images are not spatially aligned, the performance of the RGB and sonar modalities are evaluated separately.

In addition, due to the difference in perceptual range between cameras and sonar, the target may appear in only one of the modal images for paired RGB and sonar images. Therefore, the RGB-S tracker must discriminate whether the target is present in the current frame of each modality. Specifically, we set the confidence threshold of the output bounding box to $0.5\in[0,1]$. If the maximum response value of the RGB-S tracker is less than 0.5, only [0,0,0,0] is output to indicate that there is no target in the current frame. For the trackers with end-to-end prediction heads and confidence levels not between $[0,1]$, we assume that their confidence score is equal to 1.

%% file: figures/fig_rgbs50_benchmark.tex
\begin{figure*}
	\centering
        \includegraphics[width=18cm]{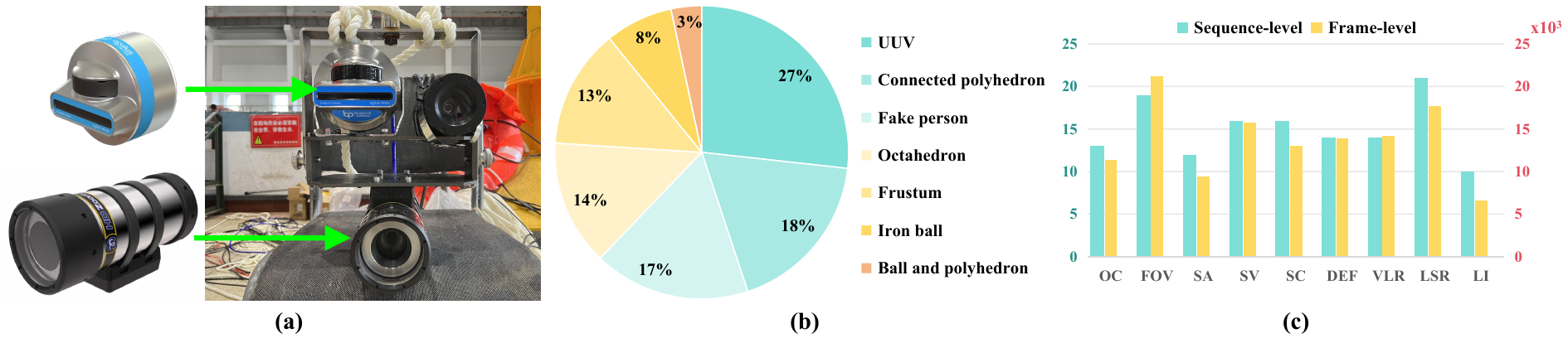}
	\caption{Main features of the proposed dataset. (a) The multi-sensor platform we use for data collection. (b) Quantity distribution of 7 object categories (name and percentage). (c) Statistics on frame-level and sequence-level attributes.}
	\label{fig: benckmark}
\end{figure*}

%% file: tables/attribute.tex
\begin{table}[]
\caption{List and description of 9 attributes in RGBS50.}
\label{table attribute}
\renewcommand{\arraystretch}{1.3}
\centering
\begin{tabular}{l|l}
\hline
\textbf{Attribute} & \textbf{Definition} \\ \hline
\textbf{OC}        & Occlusion - the target is occluded in the RGB images.    \\
\textbf{FOV}       & Full Out-of-View - the target has moved out of the field \\
                   & of 256 view in both modalities.                          \\ 
\textbf{SA}        & Similar Appearance - the target a has similar appearance \\
                   & to the surrounding objects in RGB images.                \\
\textbf{SV}        & Scale Variation - the scale change rate of the bounding box \\
                   & exceeds the range of [0.5, 2].                                       \\
\textbf{SC}        & Sonar Crossover - the target has similar sonar reflection\\
                   & value with surrounding objects.                          \\
\textbf{DEF}       & Deformation - the target exhibits significant shape changes \\
                   & in sonar images.                                     \\
\textbf{VLR}       & Visual Low Resolution - the target has low resolution in \\
                   & RGB images.                                              \\
\textbf{LI}        & Low Illumination - the illumination of RGB images is low.\\
\textbf{LSR}       & Low Sonar Reflection - the target has a low luminance    \\
                   & value in grayscale sonar images.                         \\ \hline
\end{tabular}
\end{table}

%% file: figures/fig_rgbs50_examples.tex
\begin{figure*}
	\centering
        \includegraphics[width=17cm]{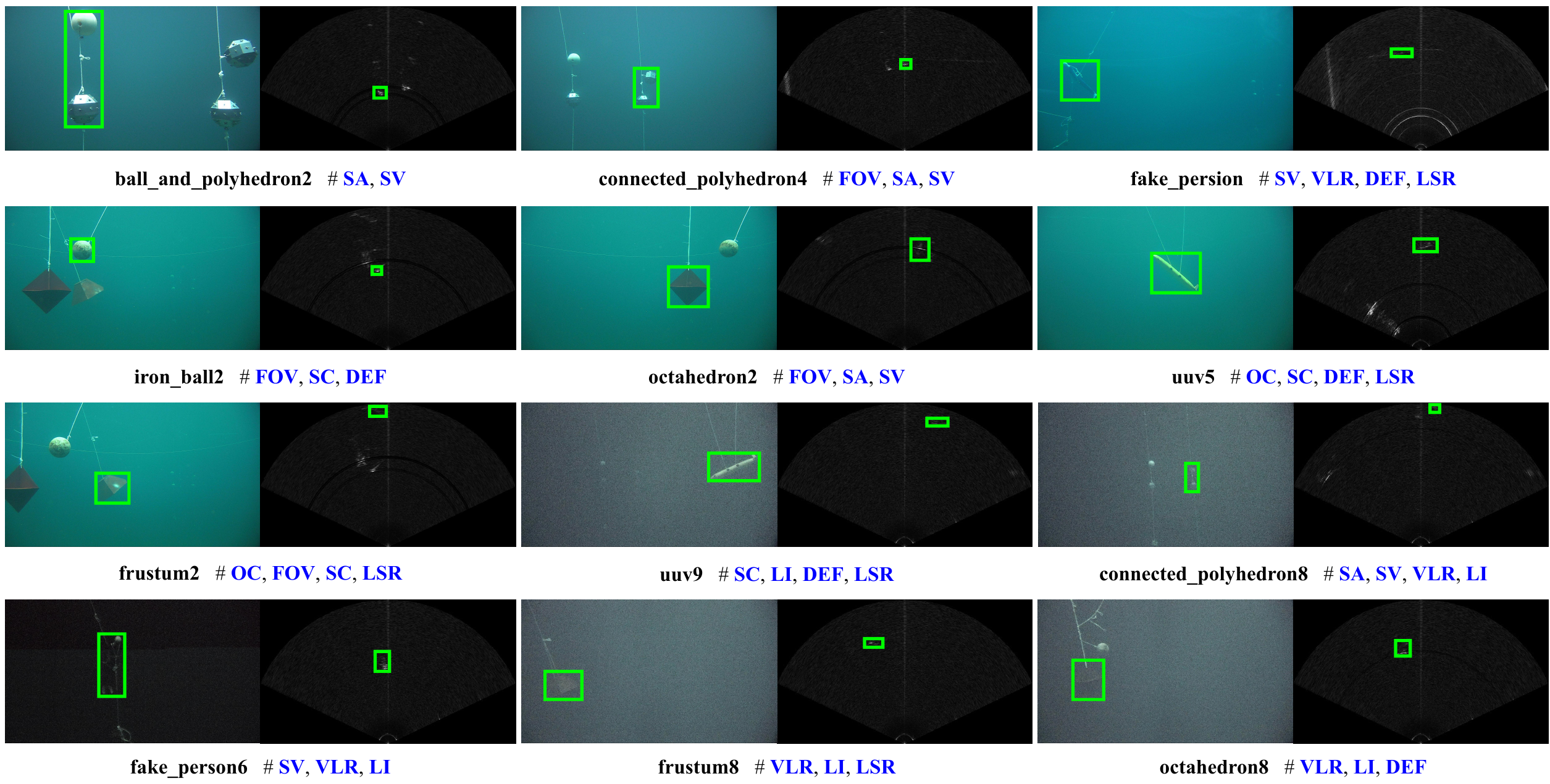}
	\caption{Sample RGB and sonar sequences with annotated attributes from our RGBS50 dataset. The \textbf{\color{black}{black}} font represents the sequence name. The \textbf{\color{blue}{blue}} font represents the annotated attributes.}
	\label{fig: rgbs50_examples}
\end{figure*}

%% file: tables/benchmarks_compare.tex
\begin{table}[]
\caption{Compare with the UOT datasets and multimodal datasets of the same magnitude.}
\label{table benchmark_compare}
\centering
\renewcommand{\arraystretch}{1.3}
\begin{tabular}{c|c|cccc}
\hline
\multirow{2}{*}{Task} & \multirow{2}{*}{Benchmark} & Num.   & Avg.   & Total  & Total       \\
                      &                            & seq.   & Frames & Frames & anno. \\ \hline
\multirow{3}{*}{UOT}  & UOT100 \cite{uot100}       & 106    & 702   & 74K   & 74K         \\
                      & UTB180 \cite{utb180}       & 180    & 338   & 58K   & 58K         \\
                      & VMAT \cite{vmat}           & 33     & 2,242 & 74K   & 74K         \\ \hline
RGB-T                 & GTOT \cite{gtot}           & 50     & 157   & 7.8K  & 15.6K       \\
RGB-T                 & VOT19-RGBT \cite{vot2019rgbe}& 60     & 334   & 40.2K & 80.4K       \\ \hline
RGB-D                 & PTB \cite{ptb}             & 100    & 203   & 20K   & 40K         \\
RGB-D                 & STC \cite{stc}             & 36     & 255   & 9.2K  & 18.4K       \\ \hline
RGB-S                 & RGBS50                     & 50     & 874   & 43.7K & 87.4K       \\ \hline
\end{tabular}
\end{table}

%% file: sections/4_method.tex
\input{figures/fig_rgbs_framework}

\section{Method}

The overall framework of SCANet is shown in Figure \ref{fig: framework}. We first embed and concatenate the multimodal template and search area images into tokens, and then input them into the ViT blocks of each branch. Our proposed spatial cross-attention module (SCAM) is inserted between ViT blocks to achieve multi-level interaction between RGB and sonar modalities. The features output from each branch are fed into the corresponding prediction head to predict the state of the target in each modality. The training of SCANet is implemented only on the SOT training dataset and some object detection datasets.

\subsection{Backbone}

Following the current popular design of multimodal trackers, we select SOT pre-trained ViT \cite{ostrack} as our backbone for joint feature extraction and modeling of template and search area.

The template images of RGB and sonar modalities are represented as $z_{r}, z_{s}\in R^{H_{z}\times W_{z}\times 3}$ and the search area images of two modalities are represented as $x_{r}, x_{s}\in R^{H_{x}\times W_{x}\times 3}$. They are first embedded as tokens $Z_{r}, Z_{s}\in R^{N_{z}\times C}$ and $X_{r}, X_{s}\in R^{N_{x}\times c}$ as same in \cite{ostrack}. $N_{i}=H_{i}W_{i}/S, i\in\{z,x\}$, where $S$ is the stride of patch embedding. $C$ is the channel dimension. We concatenate the tokens of the template and search area to obtain $H_{i}=[Z_{i};X_{i}], i\in\{r,s\}$. Then we fed $H_{z}$ and $H_{x}$ into transformer blocks for joint feature extraction and relationship modeling, represented as:
\begin{equation}
\begin{split}
A_{i} &= \text{Softmax}(\frac{QK^{T}}{\sqrt{C}})V=\text{Softmax}(\frac{[X_{i};Z_{i}][X_{i};Z_{i}]^{T}}{\sqrt{C}})V \\
      & =\text{Softmax}(\frac{[X_{i}X_{i}^{T},X_{i}Z_{i}^{T};Z_{i}X_{i}^{T},Z_{i}Z_{i}^{T}]}{\sqrt{C}})V, i\in\{r,s\}
\end{split}
\end{equation}
where Q, K, V denotes query, key and values matrices, respectively. 

\subsection{Spatial Cross-Attention Module}

The proposed SCAM aims at effective cross-modal interaction of spatially unaligned RGB features with sonar features. The SCAM consists of a spatial cross-attention layer and two independent global integration modules (GIM). The operation of our SCAM module is described as below:

The inputs to the SCAM are $H_{r}$ and $H_{s}$. First, we calculate the QKV matrices for two modalities. The Query matrix of the RGB modality is represented as $Q_{r}=H_{r}\in R^{N\times C}$, where $N=N_{z}+N_{x}$. The Key matrix and the Value matrix are obtained by $K_{r}, V_{r}=\text{Split}(\text{Linear}(H_{r}))$, where $\text{Linear}$ represents a linear layer without bias, it maps the feature from $R^{C}$ to $R^{2C}$. $\text{Split}$ maps the feature from $R^{2C}$ to two $R^{C}$. The sonar $Q_{s}, K_{s}, V_{s}$ are obtained in the same way. 

Then we calculate spatial attention maps of two modalities. As shown in Figure \ref{fig: rgbs_sca_attn}. The spatial attention map $QK^{T}\in R^{N\times N}$ models the correspondence between each spatial patch in the template and the search area in two modalities, which allows the model to associate the same target at different positions in two images through similar semantic features.

In addition, the background, which is significantly different in the two modal images, interferes with the model's association of cross-modal features, so we introduce a ReLU activation function that zeroes out and suppresses the background interference, leaving only the key features that establish the association. These key features are then added to the original features to ensure that the features for each modality are correctly augmented. The process is described as
\begin{equation}
\begin{split}
H_{r}^{attn} &= \text{ReLU}(\frac{Q_{r}K_{s}^{T}}{\sqrt{C}})V_{r}+H_{r}\\
H_{s}^{attn} &= \text{ReLU}(\frac{Q_{s}K_{r}^{T}}{\sqrt{C}})V_{s}+H_{s}
\end{split}
\end{equation}
where $\text{ReLU}$ represents the ReLU function.

Finally, we use two independent global integration modules (GIM) to globally integrate the features of each branch separately. The GIM module consists of two linear layers with a GeLU activation function in the middle. The process is described as
\begin{equation}
\begin{split}
H_{r}^{cross}&= \text{GIM}_{r}(H_{r}^{attn})+H_{r} \\
H_{s}^{cross}&= \text{GIM}_{s}(H_{s}^{attn})+H_{s}
\end{split}
\end{equation}
where $H_{r}^{cross}$ and $H_{r}^{cross}$ are the output features of the SCAM. $\text{GIM}_{r}$ and $\text{GIM}_{s}$ represents the GIM module of different brench.

\subsection{Head and Loss}

Since the RGB and sonar modalities are not spatially aligned, we use two independent prediction heads to predict the target state in the RGB and sonar modalities, respectively. Following the design of \cite{ostrack}, our head model adopts a fully convolutional prediction head as \cite{centerhead}. Two prediction heads have the same structure but do not share any parameters. During training, they inherit the same initialization parameters from the pre-trained model.

For each prediction head, we used the focal loss \cite{weightfocalloss} to train the classification branch, while we used the l1 loss and GIoU loss \cite{giou} to train the box regression branch. The total training loss is shown as
\begin{equation}
\begin{split}
L_{total}= & L_{cls}^{rgb} + L_{cls}^{sonar} + \lambda_{iou}(L_{iou}^{rgb} + L_{iou}^{sonar}) + \\
           & \lambda_{l_{1}}(L_{1}^{rgb} + L_{1}^{sonar})
\end{split}
\end{equation}
where $\lambda_{iou}=2$ and $\lambda_{l_{1}}=5$ as same in \cite{ostrack}.

\input{figures/fig_rgbs_sca}
\input{figures/fig_rgbs_srst}

\subsection{SOT data-based training}

Due to the lack of paired RGB-S data, we propose a SOT data-based RGB-S simulation training method (SRST). It aims to achieve simulation training of the RGB-S tracker by converting the training images of SOT into a kind of sonar-like image. 

First, we use the appearance features of the same target in different spatiotemporal contexts in the SOT sequence to simulate spatial misalignment in RGB-S. Specifically, we collect two paired template and search area images in an SOT sequence, and input the two pairs of images into the RGB and sonar branches, respectively.
Second, the sonar determines the brightness of a target pixel point in the sonar image by the acoustic reflection intensity of the target. Thus, a grayscale sonar image has similarities to a gray-scale saliency image. Inspired by this analysis, we transform the input image of the sonar branch from RGB images to gray-scale saliency images (the saliency detection method is \cite{saliency}) to simulate the semantic structure of RGB-S images. The qualitative comparison between simulated training data generated by SRST and real RGB-S data is shown in Figure \ref{fig: rgbs_srst}. Although there are some differences between the pseudo data and the real data, the pseudo data reflect some semantic structures and relationships of RGB-S images.

In addition, UOSTrack \cite{uostrack} shows that learning the semantics of underwater targets by the tracker can improve its performance in UOT. Inspired by \cite{uostrack}, we hope that SCANet can achieve better modeling by learning the semantic structure information of sonar targets. Although current sonar object detection datasets such as UATD \cite{uatd} contain a lot of sonar target information, the texture information and brightness in them are too weak for the model to learn the semantic structure of sonar targets. Therefore, we introduce a synthetic aperture radar (SAR) object detection dataset \cite{sardet} with more robust semantic features similar to sonar image information to facilitate the model's simulation learning of low-level semantic information of sonar-like targets.

Therefore, the total training set of SCANet contains the LaSOT \cite{lasot}, GOT10K \cite{got10k}, TrackingNet \cite{trackingnet}, COCO \cite{coco} and SARDet \cite{sardet} datasets. 

%% file: figures/fig_rgbs_framework.tex
\begin{figure*}
	\centering
        \includegraphics[width=18cm]{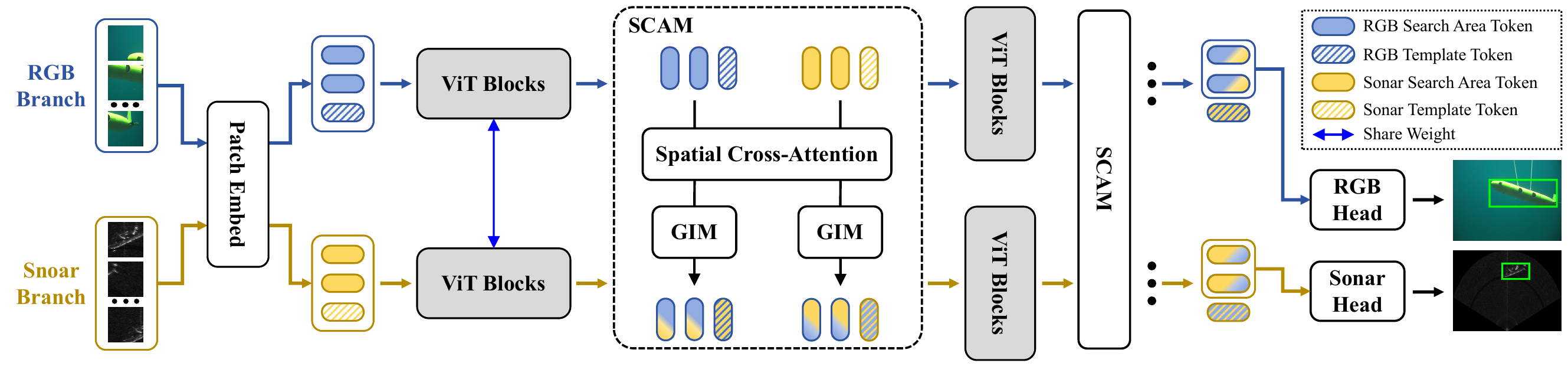}
	\caption{The overall framework of SCANet. We take the ViT pre-trained by SOT \cite{ostrack} as our baseline, and insert the proposed SCAM module into different layers in the backbone to achieve multi-level cross-modal feature interactions. Finally, the output features of each branch are fed into the corresponding prediction head to predict the state of the target in the two modalities images, respectively.}
	\label{fig: framework}
\end{figure*}

%% file: figures/fig_rgbs_sca.tex
\begin{figure}[t]
	\centering
    \includegraphics[width=8.8cm]{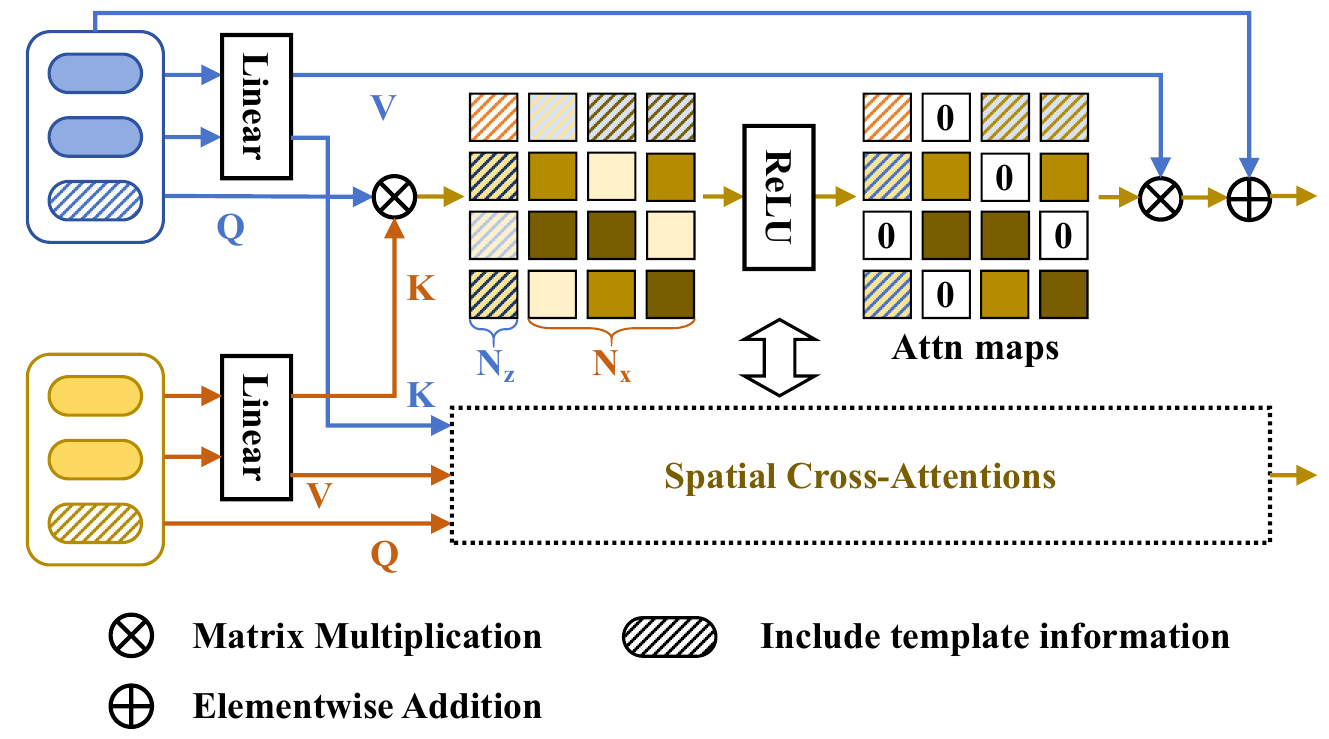}
	\caption{Illustration of the \textit{Sonar to RGB} interaction process of our proposed spatial cross-attention. The process of \textit{RGB to Sonar} is the same.}
	\label{fig: rgbs_sca_attn}
\end{figure}

%% file: figures/fig_rgbs_srst.tex
\begin{figure}
	\centering
        \includegraphics[width=8cm]{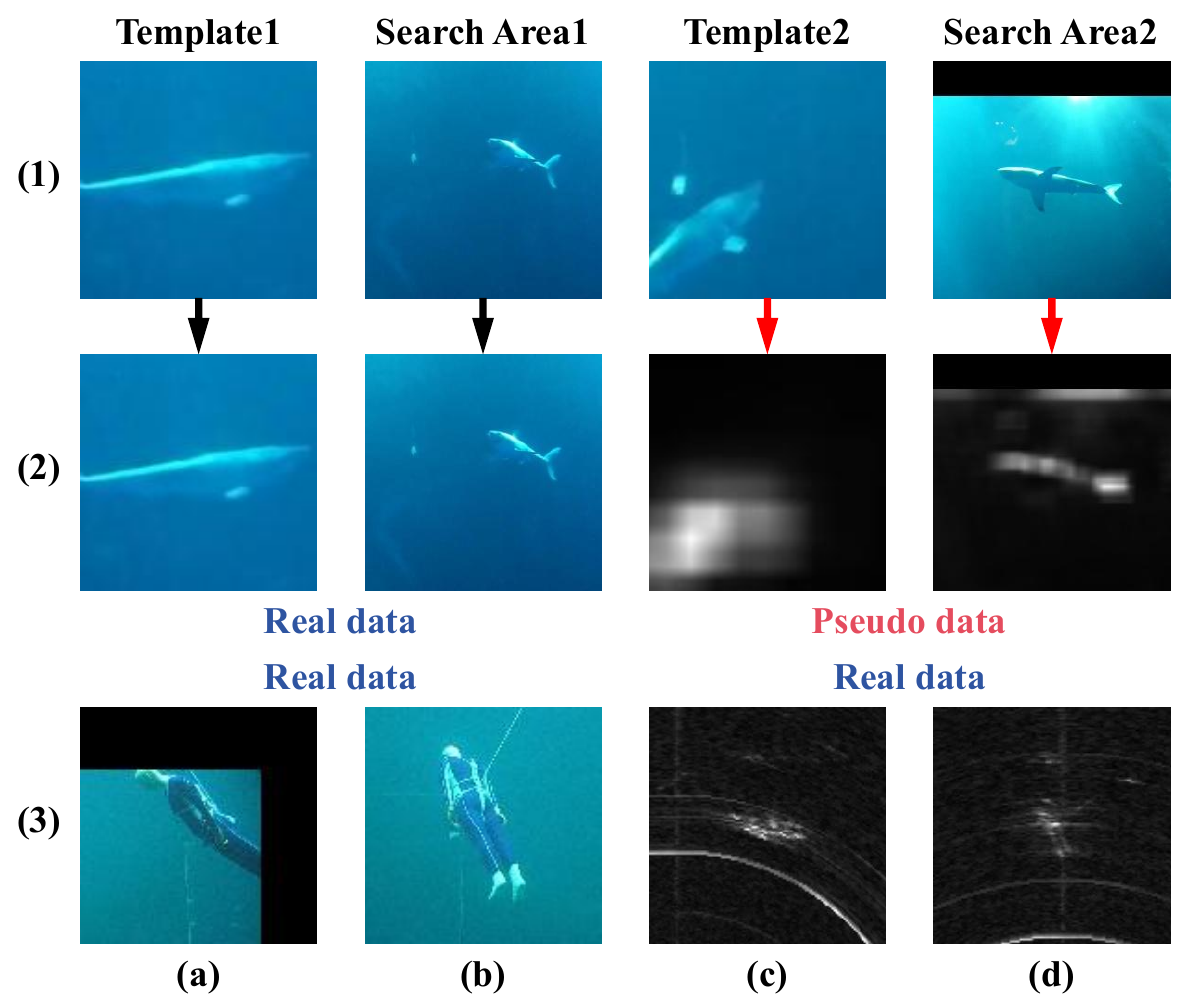}
	\caption{Comparison between our SRST training data and real RGB-sonar images. (a) The template of RGB modality. (b) The search area of RGB modality. (c) The template of sonar modality. (d) The search area of sonar modality. (1) The shark\-5 sequence of LaSOT \cite{lasot}. (2) Actual training data of our SCANet. (3) The fake\_person3 sequence of the proposed RGBS50.}

	\label{fig: rgbs_srst}
\end{figure}

%% file: sections/5_experiments.tex
\section{experiments}

\input{tables/compare_results}

\subsection{Implementation Details}
The implementation of our methods is based on Pytorch 1.13.0. The training platform consists of two NVIDIA RTX A6000 GPUs. The total number of training epochs is 10, which takes about 7.5 hours. During training, we sample 60k image pairs in each epoch, for a total batch size of 64. The total learning rate is set to $1e^{-5}$ for the backbone and two heads, and $1e^{-6}$ for the SCAM module. AdamW is used as the optimizer with a weight decay of $1e^{-4}$. The sizes of the template and search area are 128x128 and 256x256, respectively. Our proposed SCAM module is inserted into the 4-th, 7-th, and 10-th layers of the ViT backbone.

\subsection{Comparison Results}

\subsubsection{Overall Performance} 
We evaluate all the baseline methods and the proposed SCANet. The results are reported in Table \ref{table compare results}. For RGB sequences, most of these trackers have a SR score of only 20-40\%. This is due to the fact that these trackers do not effectively distinguish the presence or absence of the target within the current scene. For sonar sequences, most of the trackers have a success rate of only 30-45\%. The plots of SR, PR, and NPR are shown in Figure \ref{fig: rgbs_plots}. Overall, RGBS50 is a challenging benchmark for the baseline trackers.

SCANet achieves state-of-the-art performance. Compared to transformer-based SOT trackers, SCANet outperforms OSTrack256 \cite{ostrack} by 2.1\% in RP, 2.1\% in NPR and 1.9\% in SR for the RGB modality, respectively. For the sonar modality, SCANet has exciting results, improving the baseline OSTrack256 \cite{ostrack} by 11.6\% in PR, 10.5\% in NPR, and 9.4\% in SR, respectively. Compared to ToMP50 \cite{tomp}, which has the second best SR in the sonar modality, SCANet outperforms it by 2.8\% in SR. Compared to ODTrack \cite{ostrack}, which has similar tracking performance to our method in the RGB modality, SCANet outperforms it by 3.7\% in SR and 3.8\% in PR in the sonar sequence, respectively. In addition, although HIPTrack \cite{hiptrack} has better performance than SCANet on RGB sequences, this is due to the fact that our baseline is OSTrack \cite{ostrack}, which limits our model's feature discrimination on RGB images. In addition, SCANet outperforms HIPTrack \cite{hiptrack} by 9.8\% in SR on sonar sequences.
The results show that the interaction between the RGB modality and the sonar modality improves the tracking performance of both modalities, especially for the sonar modality, which significantly improves its performance. This further confirms the significant advantage of RGB-S tracking.

\input{figures/fig_pr_sr_plots}

\subsubsection{Attribute-based Performance}
\input{tables/attribute_results}

\input{figures/fig_rgbs_radar_sr}

\input{figures/fig_rgbs_radar_npr}

\input{tables/ablation_scam}
\input{tables/ablation_srst}

\input{tables/ablation_layers}

We first evaluate the attribute based performance of 9 representative trackers (SiamRPN++ \cite{siamrpn++}, SiamBAN-ACM \cite{siambanacm}, DiMP \cite{dimp}, ToMP \cite{tomp}, OSTrack \cite{ostrack}, ARTrackSeq \cite{artrackv2}, SeqTrack \cite{seqtrack}, ODTrack \cite{odtrack}). The results of SR and NPR are shown in Table \ref{table attribute_sr} and Table \ref{table attribute_npr}, respectively. Our proposed SCANet outperforms the other 9 trackers in most attributes. SCANet achieves state-of-the-art performance in FOV and LSR attributes. This shows that the interaction between RGB and sonar modalities effectively improves the discriminative ability of the model when there is insufficient semantic information in sonar images.
In addition, SCANet ranks in the top three for attributes such as SV and VLR, etc., demonstrating that cross-modal interaction is also equally beneficial for general tracking challenges. For the SA and DEF attributes, the performance of the SCANet is still lower than that of the single modality tracker ToMP \cite{tomp}, although the performance of the SCANet is significantly improved with respect to its baseline OSTrack \cite{ostrack}. This may be due to the fact that our model relies only on the initial template for matching, and may mismatch if there are objects of similar appearance around the target or if the appearance of the target changes significantly.

\input{figures/fig_rgbs_heatmap}

In addition, we show the success score radar maps of three representative trackers (SCANet, OSTrack \cite{ostrack}, ODTrack \cite{odtrack}) in Figure \ref{fig: radar_sr}. The proposed SCANet outperforms the baseline OSTrack \cite{ostrack} in most attributes. In particular, it improves the SR performance of OSTrack \cite{ostrack} on the FOV, SA, and SV attributes in RGB sequences. In sonar sequences, cross-modal feature interaction significantly improves the SR performance of SCANet on all attributes. However, the SR performance of SCANet is insufficient on the VLR and LI attributes in the RGB modality and on the DEF, SC, and OC attributes in the sonar modality.

The radar maps of the normalized precision score are shown in Figure \ref{fig: radar_npr}. The proposed SCANet significantly improves the NPR performance of the baseline OSTrack \cite{ostrack} on most attributes. Compared to itself, SCANet has shortcomings on SA, VLR, and LI attributes in RGB sequences. For the sonar modality, although SCANet has balanced PR performance across different attributes, there are still shortcomings in the performance of each attribute. 
 
\subsection{Ablation Studies}

To evaluate the effect of different components on SCANet, we perform the following ablation experiments on the proposed RGBS50. The evaluation metrics are PR, NPR, and SR.

\subsubsection{Ablation of SCAM}
We evaluate the contributions of each component of the SCAM. The results are reported in Table \ref{table ablation scam}. The \textit{SCA} represents the spatial cross-attention layer with the softmax function. The \textit{ReLU} represents the replacement of the softmax with the ReLU function. In the sonar sequences, the SCA improves the SR by 5.1\%, the PR by 6.3\% and the NPR by 4.7\%. Using the ReLU function results in an improvement of 2.1\% for SR, 3.9\% for PR and 3.3\% for NPR in sonar sequences. The GIM achieves 2.2\%, 1.4\% and 2.5\% in terms of SR, PR and NPR in sonar sequences, respectively. They also provide performance improvements in the RGB modality. Overall, all three components play an important role in cross-modal feature interaction.

\subsubsection{Ablation of SRST}
We evaluate the contributions of each component of the SRST. The results are reported in Table \ref{table ablation srst}. The \textit{SOT} represents the training data of the original SOT model. The \textit{ToS} represents the conversion of the RGB images of the sonar branches into the corresponding saliency images.  The \textit{WSD} represents the additional use of the SARDet dataset. The SOT improves the SR by 1.0\% and 0.5\%, the PR by 0.3\% and 0.9\% in the RGB and sonar sequences, respectively. This shows that the RGB-S tracker can be trained using only SOT training data and its performance is slightly improved. The ToS improves the SR by 0.3\% and 6.9\%, the PR by 0.6\% and 9.6\% in the RGB and sonar sequences, respectively. The WSD improves the SR by 0.6\% and 2.0\%, the PR by 1.2\% and 1.1\% in the RGB and sonar sequences, respectively. Overall, although the SCANet does not use any RGB-S data for training, the SRST succeeds in making the modal learn the semantic structure of RGB-S-like images and achieves performance improvements on the RGB-S tracking benchmark.

\subsubsection{Ablation of Insert Layers}
We explore the impact of inserting the SCAM module into different layers on model performance. The results are reported in Table \ref{table ablation layers}. When the proposed SCAM module is inserted in the 4th layer, significant performance improvements are achieved in the sonar sequences. Specifically, it outperforms the baseline by 8.0\% in SR, 10.6\% in PR, and 8.2\% in NPR, respectively. We further improved the performance of the model by inserting the SCAM module in the 7th layer. Specifically, it improves the SR by 1.3\% and 0.5\%, the PR by 1.9\% and 0.5\%, the NPR by 2.6\% and 0.4\% on RGB and sonar sequences, respectively. In addition, inserting the SCAM module in the 10th layer results in an improvement of 0.8\% and 0.7\% for SR, 1.6\% and 0.5\% for PR, 0.9\% and 1.9\% for NPR on RGB and sonar sequences, respectively. Overall, the insertion of the SCAM modules in layers 4, 7, and 10 of the backbone effectively achieves multi-level cross-modal feature interaction.

\subsection{Visualization}
\subsubsection{Heatmap} 

\input{figures/fig_rgb_trackingresults}
We further explore how SCANet achieves performance improvements by interacting the RGB modality with the sonar modality. As shown in Figure \ref{fig: rgbs_heatmap}, in the fake\_persion2 sequence, for low-resolution underwater images and low-reflection-intensity sonar images, the proposed SCANet achieves better single-modality feature modeling through cross-modal feature interaction than the baseline SOT tracker \cite{ostrack}, In addition, at frame 168 (Figure \ref{fig: rgbs_heatmap} (1)(j)), when the sonar target shows significant shape changes, OSTrack \cite{ostrack} focuses on the target's edges and background interference (Figure \ref{fig: rgbs_heatmap} (1)(k)), and SCANet focuses on the target's core features (Figure \ref{fig: rgbs_heatmap} (1)(i)). In the uuv2 sequence, our CSANet achieves richer feature representations for the typical underwater vehicle target in the RGB and sonar modalities, respectively. In the octahedron1 sequence, cross-modal information sharing helps the tracker overcome the interference of acoustic shadows in sonar images. In the frustum1 sequence, our model uses the rich visual information to compensate for the discriminative information loss caused by the low reflection intensity in the sonar modality. When the target in a sonar image is close to the edge of the image, OSTrack \cite{ostrack} focuses on a limited number of target edges, while SCANet focuses on more target information. Overall, the interaction of RGB and sonar information improves the feature modeling of each modality. 

\subsubsection{Tracking Results}
We show some results of our SCANet and other state-of-the-art trackers on four representative sequences from the proposed dataset. As shown in Figure \ref{fig: trackingresults}, in the fake\_person4 sequence, when only the sonar image contains targets, SCANet still achieves better localization and scale estimation of sonar targets through multimodal template feature interactions. In the uuv9 sequence, at frame 325, the trackers accurately track the target in the RGB image, but when the targets disappear from the RGB image at frame 645, only our SCANet discriminates that there are no targets in the image. In the octahedron2 sequence, our proposed SCANet has better robustness. For example, in frames 567 and 657, the OSTrack \cite{ostrack} incorrectly determines that the target does not exist in the current scene. The bounding box predicted by ODTrack \cite{odtrack} covers the entire sonar image. Although SiamBAN-ACM \cite{siambanacm} does not lose the target, its bounding box prediction is inaccurate compared to SCANet. When the target has lower sonar reflectance values, i.e., the target has lower brightness and weaker texture information in the sonar image, the SCANet can track the target accurately and consistently compared to other trackers. Overall, the improvement of our proposed SCANet in RGB sequences is not significant, which corresponds to the fact that SCANet only outperforms ODTrack by 0.5\% in SR in Section 4.2. This is due to the simplicity of the RGB sequences in our dataset compared to the SOT dataset. In the sonar sequence, SCANet achieves better feature discrimination.

%% file: tables/compare_results.tex
\begin{table*}[]
\caption{Comparison results for our method and the baseline methods on the proposed dataset. The best two results are shown in \color{red}{red} \color{black}{and} \color{blue}{blue} \color{black}{fonts}.}
\label{table compare results}
\centering
\renewcommand{\arraystretch}{1.4}
\small %
\begin{tabular}{c|c|c|c|ccc|ccc}
\hline
\multirow{2}{*}{Trackers} & \multirow{2}{*}{Type} & \multirow{2}{*}{Source}    & \multirow{2}{*}{baseline}  & \multicolumn{3}{c|}{RGB}                     & \multicolumn{3}{c}{Sonar}                                          \\ \cline{5-10} 
                          &                       &            &                            & SR                   & PR                   & NPR                   & SR                   & PR                    & NPR                  \\ \hline
SiamRPN \cite{siamrpn}    & SOT                   & CVPR2018   & Siamese                    & 59.3                 & 57.6                 & 62.7                  & 35.7                 & 50.0                 & 42.6                 \\
SiamRPN++ \cite{siamrpn++}& SOT                   & CVPR2019   & Siamese                    & 55.7                 & 55.3                 & 62.2                  & 46.3                 & 71.0                 & 56.2                 \\
SiamCAR \cite{siamcar}    & SOT                   & CVPR2020   & Siamese                    & 22.0                 & 19.0                 & 24.5                  & 36.9                 & 64.0                 & 45.6                 \\
SiamBAN \cite{siamban}    & SOT                   & CVPR2020   & Siamese                    & 30.8                 & 31.5                 & 33.9                  & 41.6                 & 65.8                 & 51.6                 \\
SiamBAN-ACM \cite{siambanacm}& SOT                & CVPR2021   & Siamese                    & 58.3                 & 60.4                 & 63.0                  & 43.8                 & 69.4                 & 58.1                 \\ \hline\
DiMP18 \cite{dimp}        & SOT                   & CVPR2019   & DCF                        & 24.0                 & 19.3                 & 25.4                  & 41.9                 & 67.3                 & 51.4                 \\
DiMP50 \cite{dimp}        & SOT                   & CVPR2019   & DCF                        & 25.7                 & 21.0                 & 28.0                  & 46.4                 & \color{red}{72.3}    & 56.9                 \\
PrDiMP18 \cite{prdimp}    & SOT                   & CVPR2020   & DCF                        & 22.4                 & 19.6                 & 23.5                  & 31.6                 & 54.9                 & 35.6                 \\
PrDiMP50 \cite{prdimp}    & SOT                   & CVPR2020   & DCF                        & 27.7                 & 22.6                 & 30.5                  & 36.0                 & 55.7                 & 42.6                 \\
KeepTrack \cite{keeptrack}& SOT                   & ICCV2021   & DCF                        & 33.5                 & 29.8                 & 38.2                  & 39.6                 & 61.5                 & 45.9                 \\ \hline
TrDiMP \cite{trdimp}      & SOT                   & CVPR2021   & DCF+Transformer            & 27.2                 & 22.7                 & 30.9                  & 37.3                 & 58.8                 & 43.5                 \\ 
ToMP50 \cite{tomp}        & SOT                   & ICCV2022   & DCF+Transformer            & 32.7                 & 32.7                 & 37.5                  & \color{blue}{47.5}   & \color{blue}{70.8}   & \color{blue}{61.1}   \\ \hline
Stark-S50 \cite{stark}    & SOT                   & ICCV2021   & Transformer                & 35.0                 & 35.8                 & 40.0                  & 40.2                 & 60.5                 & 50.1                 \\
Stark-ST50 \cite{stark}   & SOT                   & ICCV2021   & Transformer                & 35.2                 & 35.4                 & 39.4                  & 41.1                 & 62.3                 & 51.4                 \\
Stark-ST101 \cite{stark}  & SOT                   & ICCV2021   & Transformer                & 35.2                 & 35.2                 & 39.9                  & 40.2                 & 60.5                 & 50.1                 \\
TransT \cite{transt}      & SOT                   & CVPR2021   & Transformer                & 28.0                 & 29.0                 & 31.9                  & 42.9                 & 65.1                 & 50.5                 \\
SLT-TransT \cite{slttranst}& SOT                  & ECCV2022   & Transformer                & 27.9                 & 28.9                 & 31.7                  & 42.9                 & 65.1                 & 50.6                 \\
AiATrack \cite{aiatrack}  & SOT                   & ECCV2022   & Transformer                & 34.6                 & 34.7                 & 40.9                  & 43.0                 & 67.3                 & 51.4                 \\
CSWinTT \cite{cswintt}    & SOT                   & CVPR2022   & Transformer                & 33.6                 & 32.6                 & 38.9                  & 34.7                 & 51.4                 & 43.4                 \\
OSTrack256 \cite{ostrack} & SOT                   & ECCV2022   & Transformer                & 69.3                 & 73.5                 & 75.9                  & 40.9                 & 54.6                 & 51.6                 \\
OSTrack384 \cite{ostrack} & SOT                   & ECCV2022   & Transformer                & 69.5                 & 73.8                 & 75.6                  & 33.3                 & 43.4                 & 40.4                 \\ 
SeqTrack-B256\cite{seqtrack}& SOT                 & CVPR2023   & Transformer                & 38.6                 & 40.1                 & 43.9                  & 38.3                 & 59.6                 & 46.7                 \\ 
ARTrackSeq-256\cite{artrackv2}& SOT               & arXiv2024  & Transformer                & 36.6                 & 37.0                 & 39.7                  & 47.0                 & 68.7                 & 54.6                 \\ 
ODTrack-B \cite{odtrack}  & SOT                   & AAAI2024   & Transformer                & 70.7                 & 73.9                 & 77.4                  & 44.6                 & 60.4                 & 53.9                 \\
HIPTrack \cite{hiptrack}  & SOT                   & CVPR2024   & Transformer                & \color{red}{72.1}    & \color{red}{76.8}    & \color{red}{78.3}     & 40.5                 & 53.0                 & 49.0                 \\ \hline
SCANet                    & RGB-S                 & -          & Transformer                & \color{blue}{71.2}   & \color{blue}{75.6}   & \color{blue}{78.0}    & \color{red}{50.3}    & 66.2                 & \color{red}{62.1}   \\ \hline
\end{tabular}
\end{table*}

%% file: figures/fig_pr_sr_plots.tex
\begin{figure*}[] %
\begin{minipage}[]{0.33\linewidth} 
\centering
\includegraphics[width=6cm,height=5cm]{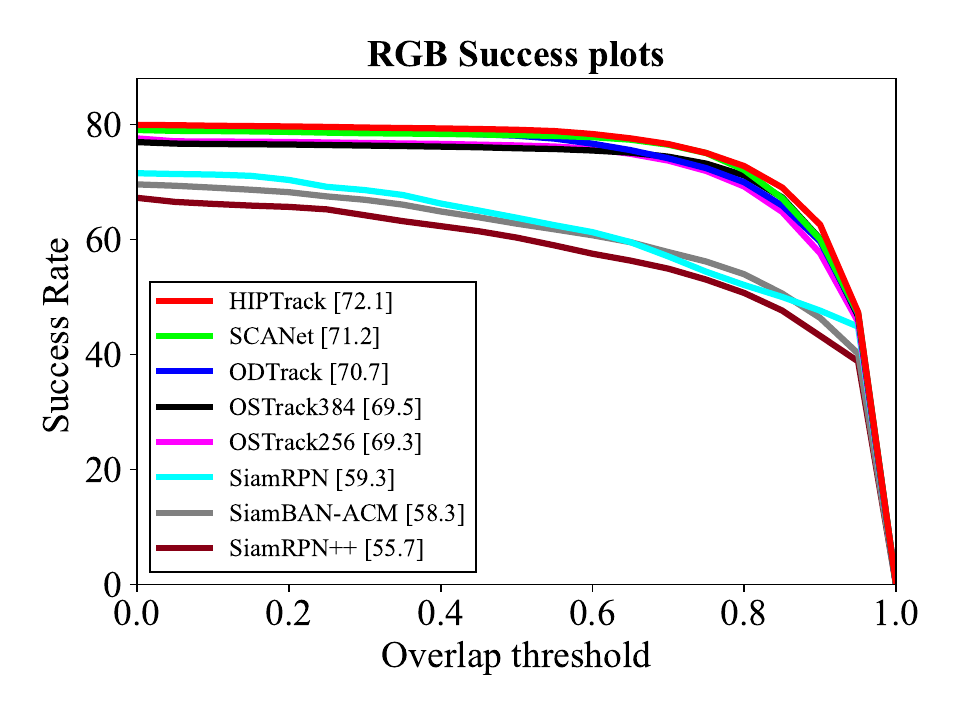} 
\end{minipage}%
\begin{minipage}[]{0.33\linewidth}
\centering
\includegraphics[width=6cm,height=5cm]{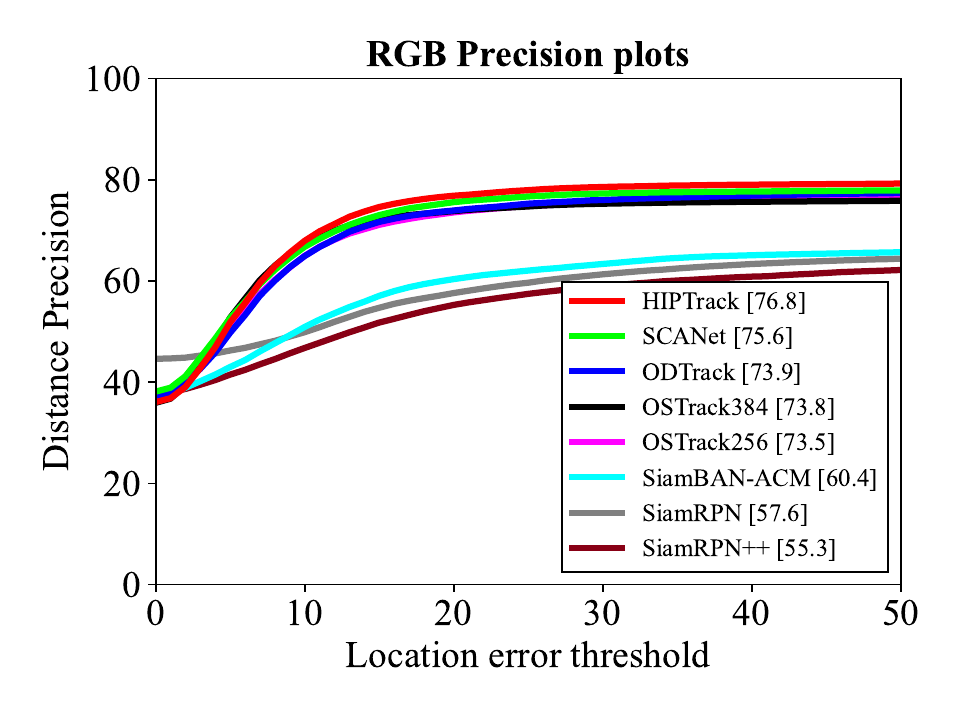}
\end{minipage}%
\begin{minipage}[]{0.33\linewidth}
\centering
\includegraphics[width=6cm,height=5cm]{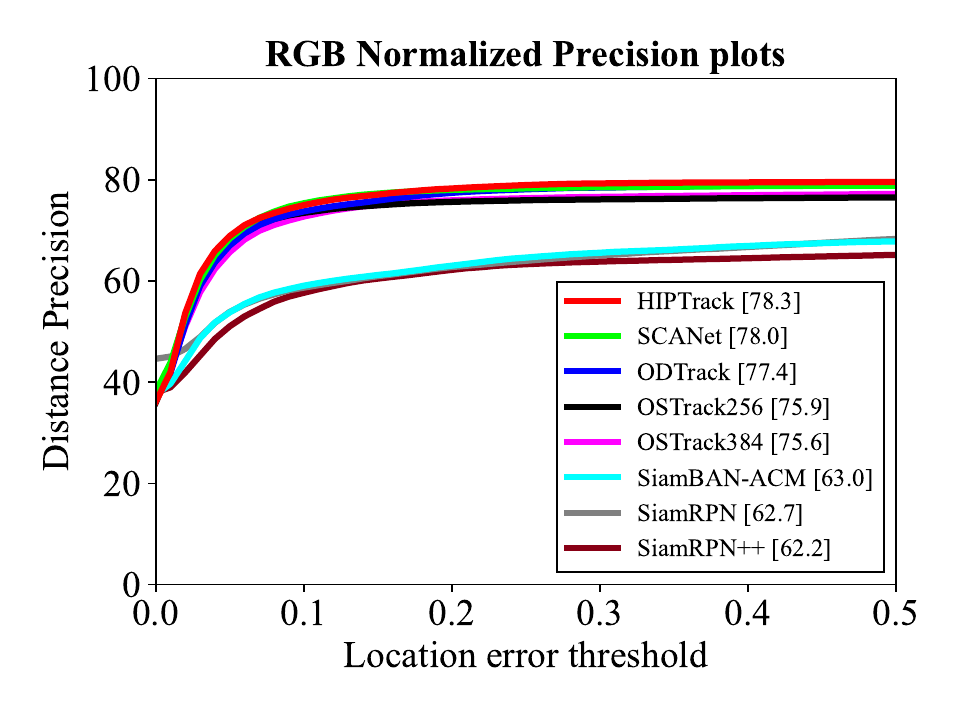}
\end{minipage}
\begin{minipage}[]{0.33\linewidth} 
\centering
\includegraphics[width=6cm,height=5cm]{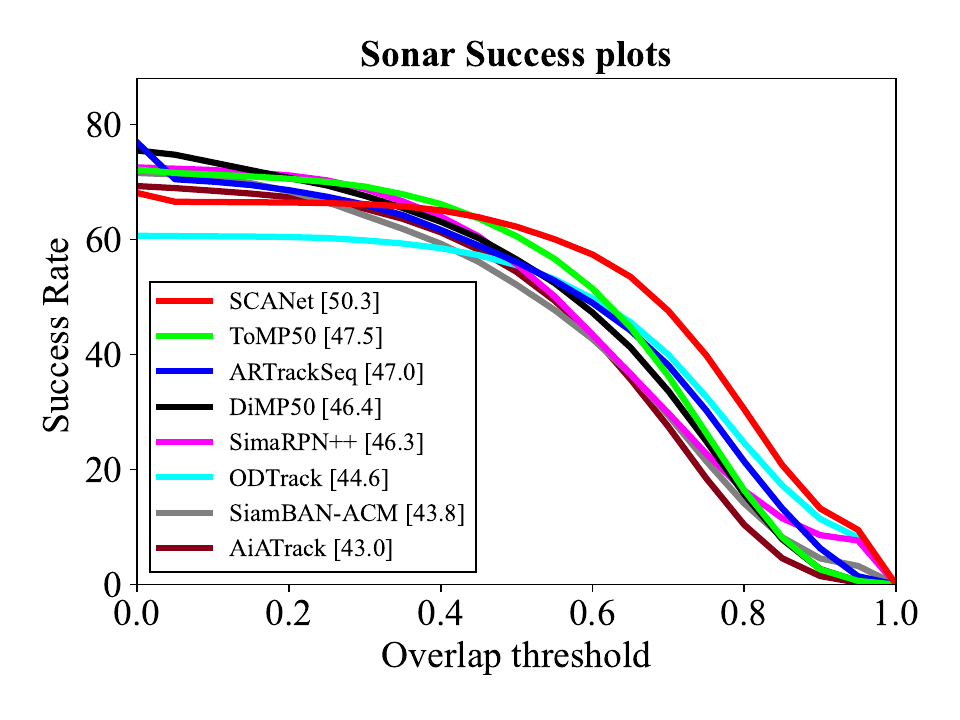} 
\end{minipage}%
\begin{minipage}[]{0.33\linewidth}
\centering
\includegraphics[width=6cm,height=5cm]{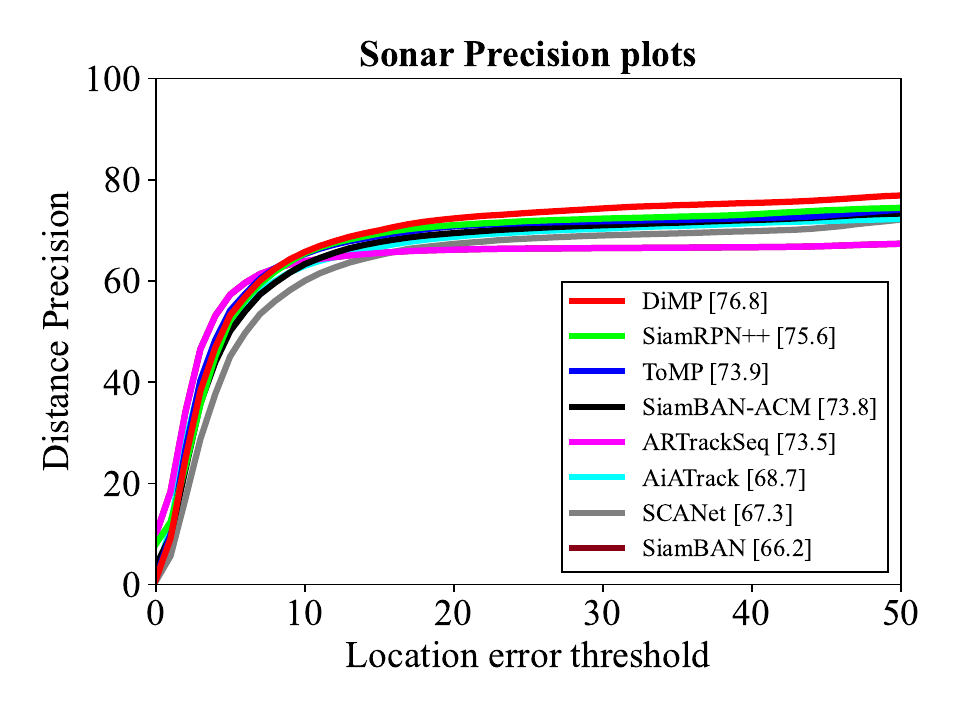}
\end{minipage}%
\begin{minipage}[]{0.33\linewidth}
\centering
\includegraphics[width=6cm,height=5cm]{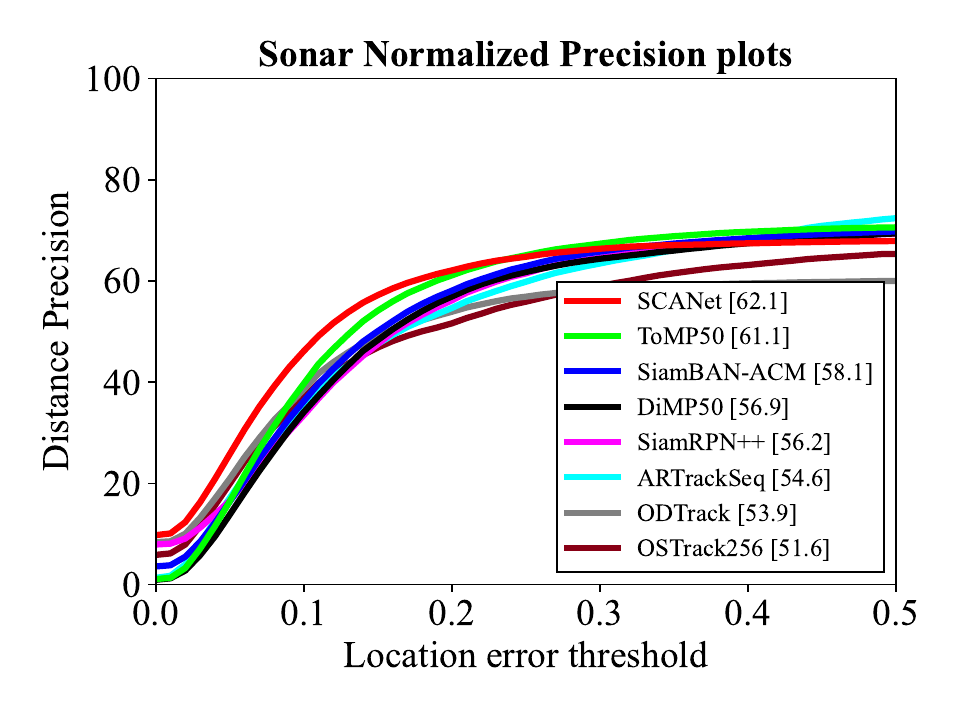}
\end{minipage}
\caption{The success plots, precision plots and normalized precision plots of the trackers (SCANet, HIPTrack \cite{hiptrack}, ODTrack \cite{odtrack}, ARTrackSeq \cite{artrackv2}, OSTrack256 \cite{ostrack}, OSTrack384 \cite{ostrack}, AiATrack \cite{aiatrack}, ToMP50 \cite{tomp}, SiamBAN-ACM\cite{siambanacm}, SiamBAN \cite{siamban}, DiMP50 \cite{dimp}, SiamRPN++ \cite{siamrpn++}, SiamRPN \cite{siamrpn}) on the proposed benchmark. In each metric, we only show the top eight trackers with the highest performance.}
\label{fig: rgbs_plots}

\end{figure*}

%% file: tables/attribute_results.tex
\begin{table*}[]
\caption{Attribute based SR of 9 trackers on RGBS50. The best three results are shown in \color{red}{red}\color{black}{,} \color{blue}{blue} \color{black}{and} \color{green}{green} \color{black}{fonts}.}
\label{table attribute_sr}
\centering
\renewcommand{\arraystretch}{1.45}
\begin{tabular}{c|ccccccccc}
\hline
\multirow{3}{*}{} & \multicolumn{9}{c}{RGB / Sonar}                                                                                                    \\ \cline{2-10} 
                  & SiamRPN++ & SiamBAN-ACM & DiMP50    & ToMP50    & OSTrack256 & ARTrackSeq-256 & SeqTrack-B25 & ODTrack-B & \multirow{2}{*}{SCANet} \\
                  & \cite{siamrpn++}& \cite{siambanacm}& \cite{dimp} & \cite{tomp} & \cite{ostrack} & \cite{artrackv2}  & \cite{seqtrack} & \cite{odtrack} &  \\ \hline
OC                & 45.2/\color{blue}{39.5} & 56.0/35.1   & 21.3/33.0 & 31.9/29.4 & \color{blue}{65.5}\color{black}{/}34.6  & 40.9/37.1      & 40.7/28.1    & \color{green}{63.8}\color{black}{/}\color{green}{38.5} & \color{red}{67.1}\color{black}{/}\color{red}{43.3 }              \\
FOV               & 58.4/\color{green}{40.4} & 64.4/32.4   & 22.1/33.4 & 28.4/\color{blue}{42.5} & \color{green}{64.8}\color{black}{/}31.4  & 28.5/34.0      & 33.5/28.8    & \color{red}{69.1}\color{black}{/}39.5 & \color{blue}{67.8}\color{black}{/}\color{red}{47.1}               \\
SA                & 57.6/\color{red}{52.7} & 57.2/49.3   & 20.2/\color{green}{51.8} & 29.2/51.5 & \color{blue}{63.2}\color{black}{/}42.4  & 25.8/\color{blue}{52.0}      & 32.0/36.7    & \color{green}{60.9}\color{black}{/}47.7 & \color{red}{63.7}\color{black}{/}51.7               \\
SV                & 56.8/\color{green}{52.5} & 59.6/49.8   & 28.2/51.8 & 35.8/\color{red}{56.2} & \color{green}{69.4}\color{black}{/}46.0  & 37.8/52.4      & 39.1/42.2    & \color{blue}{69.8}\color{black}{/}49.7 & \color{red}{70.9}\color{black}{/}\color{blue}{56.0}               \\
SC                & 62.5/40.6 & 65.4/34.6   & 23.5/\color{blue}{44.6} & 30.6/\color{red}{45.0} & \color{green}{66.7}\color{black}{/}35.6  & 36.8/43.2      & 38.0/41.1    & \color{red}{71.9}\color{black}{/}42.2 & \color{blue}{70.0}\color{black}{/}\color{green}{43.7}               \\
DEF               & 57.5/\color{blue}{43.2} & 60.0/38.6   & 26.7/\color{green}{41.0} & 31.9/\color{red}{45.0} & \color{green}{70.0}\color{black}{/}28.0  & 35.0/38.0      & 36.5/34.4    & \color{blue}{72.3}\color{black}{/}32.4 & \color{red}{74.0}\color{black}{/}39.2               \\
VLR               & 44.0/\color{red}{44.7} & 43.1/37.9   & 24.5/39.6 & 29.8/40.9 & \color{blue}{62.0}\color{black}{/}38.8  & 33.2/38.5      & 38.0/30.3    & \color{green}{60.2}\color{black}{/}\color{blue}{42.8} & \color{red}{64.1}\color{black}{/}\color{green}{42.3}               \\
LI                & 44.6/54.0 & 35.0/52.4   & 19.7/\color{green}{54.8} & 28.8/\color{blue}{56.4} & \color{red}{65.1}\color{black}{/}45.9  & 31.0/\color{red}{57.1}      & 31.2/42.4    & \color{green}{61.5}/52.5 & \color{blue}{64.0}\color{black}{/}51.1               \\
LSR               & 51.3/37.5 & 53.1/41.5   & 29.3/43.2 & 37.1/40.2 & \color{green}{72.1}\color{black}{/}44.0  & 44.7/\color{green}{46.1}      & 45.8/40.3    & \color{blue}{72.3}\color{black}{/}\color{blue}{47.2} & \color{red}{75.4}\color{black}{/}\color{red}{49.9}               \\ \hline
ALL               & 55.7/46.3 & 58.3/43.8   & 25.7/46.4 & 32.7/\color{blue}{47.5} & \color{green}{69.3}\color{black}{/}40.9  & 36.6/\color{green}{47.0}      & 38.6/38.3    & \color{blue}{70.7}\color{black}{/}44.6 & \color{red}{71.2}\color{black}{/}\color{red}{50.3}               \\ \hline
\end{tabular}
\end{table*}

\begin{table*}[]
\caption{Attribute based NPR of 9 trackers on RGBS50. The best three results are shown in \color{red}{red}\color{black}{,} \color{blue}{blue} \color{black}{and} \color{green}{green} \color{black}{fonts}.}
\label{table attribute_npr}
\centering
\renewcommand{\arraystretch}{1.45}
\begin{tabular}{c|ccccccccc}
\hline
\multirow{3}{*}{} & \multicolumn{9}{c}{RGB / Sonar}                                                                                                    \\ \cline{2-10} 
                  & SiamRPN++ & SiamBAN-ACM & DiMP50    & ToMP50    & OSTrack256 & ARTrackSeq-256 & SeqTrack-B25 & ODTrack-B & \multirow{2}{*}{SCANet} \\
                  & \cite{siamrpn++}& \cite{siambanacm}& \cite{dimp} & \cite{tomp} & \cite{ostrack} & \cite{artrackv2}  & \cite{seqtrack} & \cite{odtrack} &  \\ \hline
OC                & 50.9/\color{blue}{47.8} & 59.8/\color{green}{46.8}   & 23.5/39.9 & 34.6/38.8 & \color{blue}{73.3}\color{black}{/}44.1  & 44.9/43.0      & 47.4/36.7    & \color{green}{70.3}\color{black}{/}45.9 & \color{red}{74.3}\color{black}{/}\color{red}{55.3}              \\
FOV               & 63.7/\color{green}{48.1} & 68.7/40.7   & 24.1/41.7 & 32.5/\color{blue}{54.7} & \color{green}{69.9}\color{black}{/}39.6  & 31.0/40.1      & 38.1/34.9    & \color{red}{74.8}\color{black}{/}47.5 & \color{blue}{73.1}\color{black}{/}\color{red}{56.4}              \\
SA                & 64.7/62.5 & 58.7/\color{blue}{66.9}   & 19.4/\color{green}{65.2} & 35.5/\color{red}{67.4} & \color{blue}{69.2}\color{black}{/}52.9  & 28.4/64.5      & 37.7/45.7    & \color{green}{68.4}\color{black}{/}60.1 & \color{red}{69.9}\color{black}{/}63.4              \\
SV                & 63.5/61.9 & 62.3/\color{green}{67.3}   & 30.1/63.6 & 42.0/\color{red}{73.6} & \color{green}{75.9}\color{black}{/}57.5  & 41.1/62.1      & 45.1/52.8    & \color{blue}{77.0}\color{black}{/}62.0 & \color{red}{77.8}\color{black}{/}\color{blue}{68.5}              \\
SC                & 68.7/52.7 & 69.9/48.7   & 28.0/\color{blue}{57.5} & 33.4/\color{red}{58.9} & \color{green}{74.0}\color{black}{/}46.7  & 40.2/51.3      & 43.3/51.2    & \color{red}{78.6}\color{black}{/}50.2 & \color{blue}{77.0}\color{black}{/}\color{green}{55.3}              \\
DEF               & 62.9/\color{blue}{55.6} & 64.1/53.0   & 31.2/\color{green}{54.1} & 36.1/\color{red}{57.6} & \color{green}{75.7}\color{black}{/}37.0  & 36.6/43.5      & 40.0/41.0    & \color{blue}{77.5}\color{black}{/}37.5 & \color{red}{79.8}\color{black}{/}49.0              \\
VLR               & 49.9/\color{blue}{50.9} & 45.8/\color{red}{53.2}   & 26.1/45.6 & 35.8/47.6 & \color{blue}{68.7}\color{black}{/}47.1  & 36.5/39.9      & 44.3/32.8    & \color{green}{66.7}\color{black}{/}\color{green}{49.0} & \color{red}{71.2}\color{black}{/}\color{blue}{50.9}              \\
LI                & 50.5/63.4 & 36.7/\color{red}{73.0}   & 19.8/\color{green}{64.4} & 33.4/\color{blue}{65.5} & \color{red}{71.8}\color{black}{/}56.3  & 33.1/62.2      & 35.6/47.0    & \color{green}{67.6}\color{black}{/}61.4 & \color{blue}{71.0}\color{black}{/}62.2              \\
LSR               & 58.2/45.7 & 59.2/\color{blue}{56.0}   & 31.7/50.8 & 42.0/51.9 & \color{blue}{80.0}\color{black}{/}54.6  & 49.3/50.6      & 52.5/48.5    & \color{green}{79.6}\color{black}{/}\color{green}{55.7} & \color{red}{83.4}\color{black}{/}\color{red}{62.4}              \\ \hline
ALL               & 62.2/56.2 & 63.0/\color{green}{58.1}   & 28.0/56.9 & 37.5/\color{blue}{61.1} & \color{green}{75.9}\color{black}{/51.6}  & 39.7/54.6      & 43.9/46.7    & \color{blue}{77.4}\color{black}{/53.9} & \color{red}{78.0}\color{black}{/}\color{red}{62.1}               \\ \hline
\end{tabular}
\end{table*}

%% file: figures/fig_rgbs_radar_sr.tex
\begin{figure}[t]
    \begin{minipage}[t]{0.5\linewidth}
        \centering
        \includegraphics[width=4.4cm]{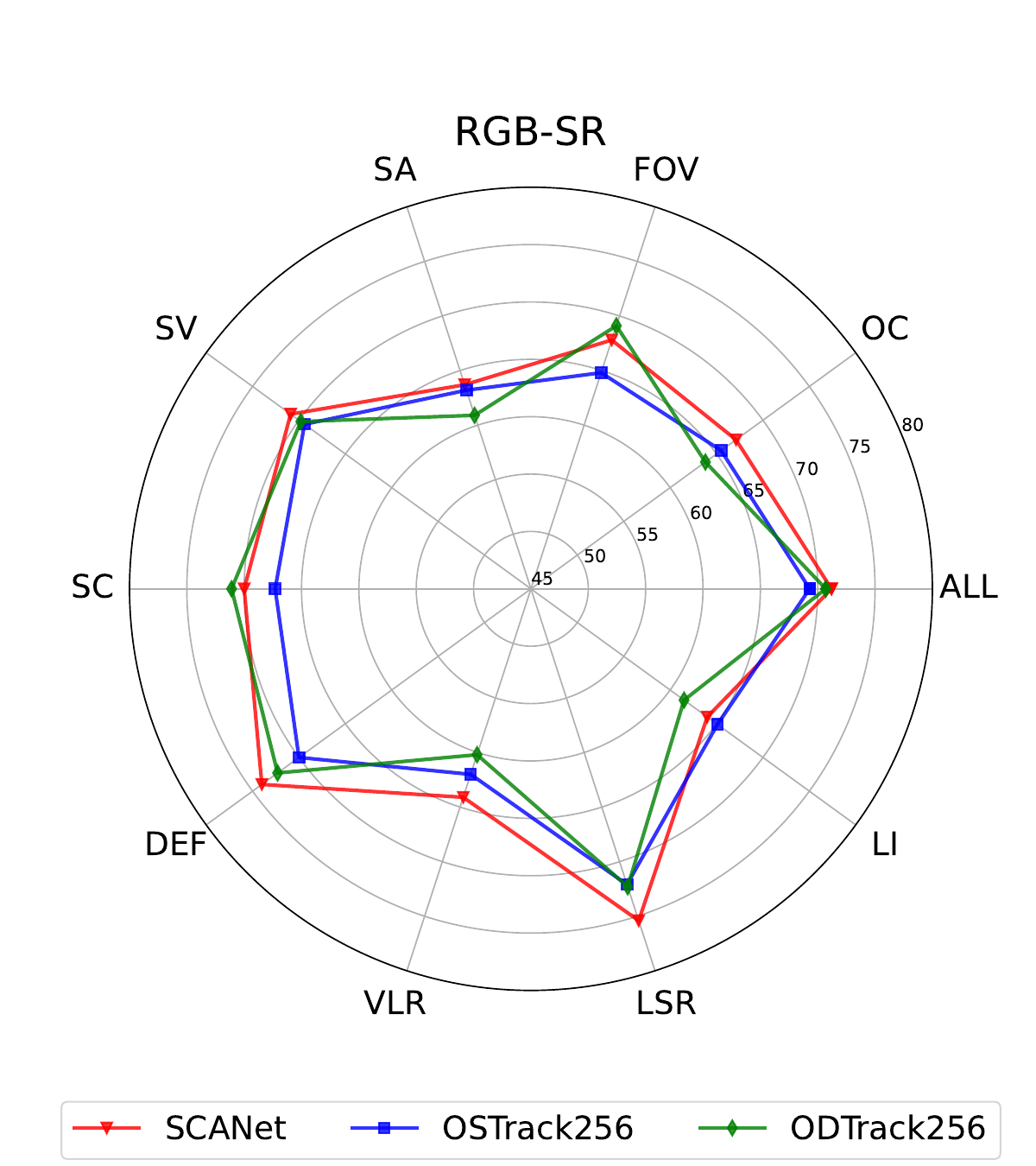}
    \end{minipage}%
    \begin{minipage}[t]{0.5\linewidth}
        \centering
        \includegraphics[width=4.4cm]{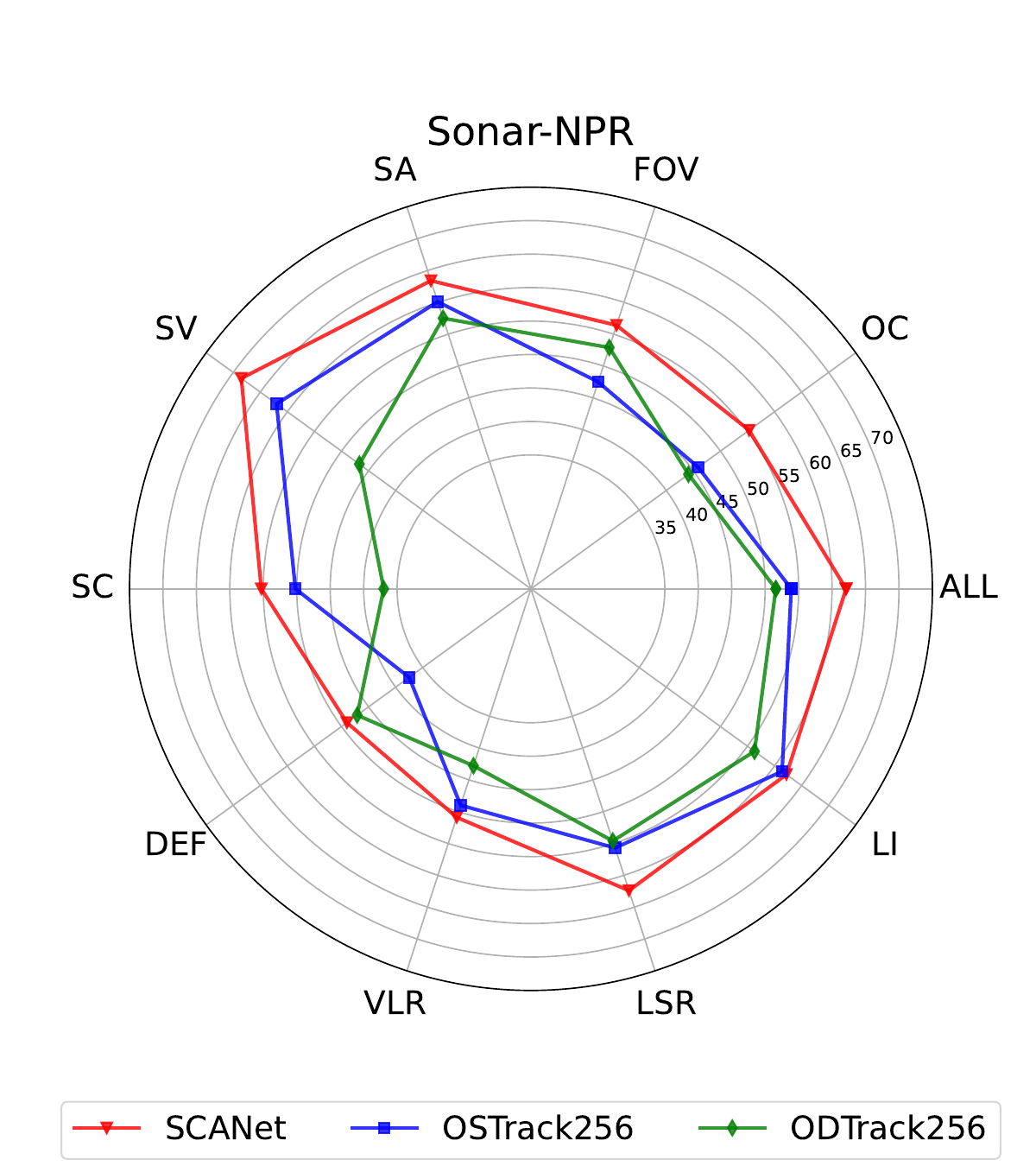}
    \end{minipage}
    \caption{The success rate scores of SCANet, OSTrack \cite{tbsi}, and ODTrack \cite{odtrack} under different attributes on the proposed benchmark. }
    \label{fig: radar_sr}
\end{figure}

%% file: figures/fig_rgbs_radar_npr.tex
\begin{figure}[t]
    \begin{minipage}[t]{0.5\linewidth}
        \centering
        \includegraphics[width=4.4cm]{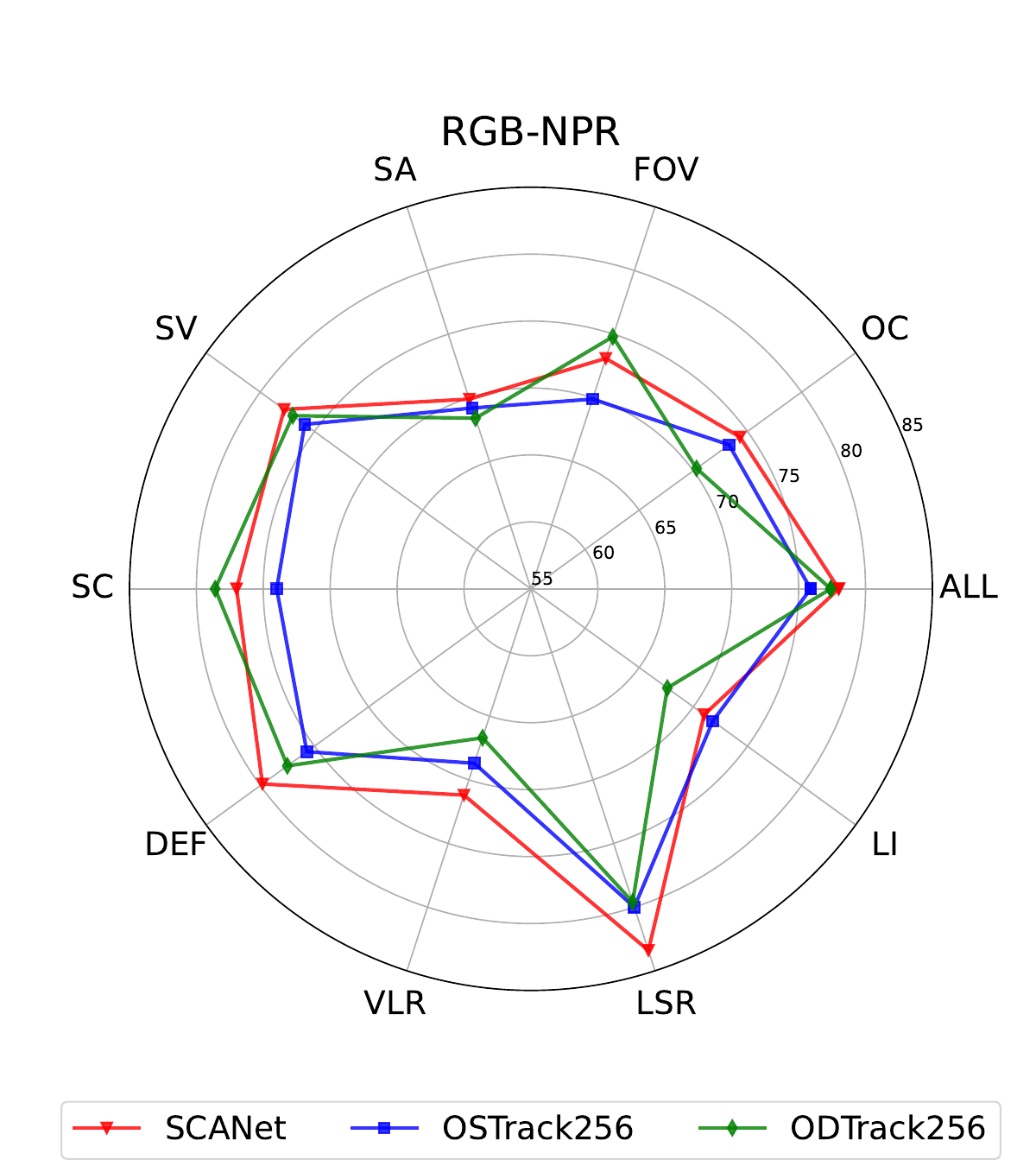}
    \end{minipage}%
    \begin{minipage}[t]{0.5\linewidth}
        \centering
        \includegraphics[width=4.4cm]{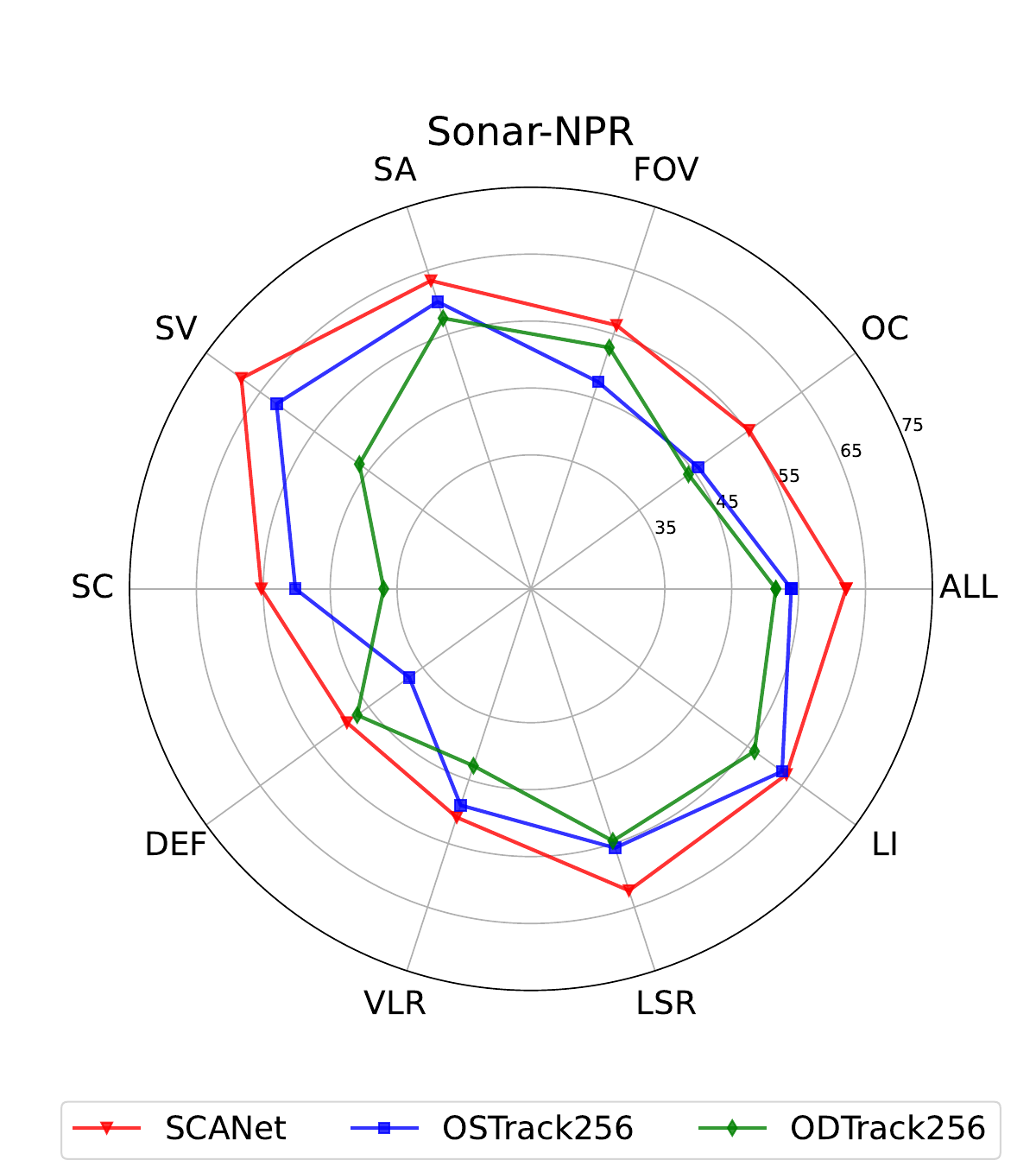}
    \end{minipage}
    \caption{The normalized precision scores of SCANet, OSTrack \cite{tbsi}, and ODTrack \cite{odtrack} under different attributes on the proposed benchmark. }
    \label{fig: radar_npr}
\end{figure}

%% file: tables/ablation_scam.tex
\begin{table}[]
\caption{Ablation of different components of SCAM.}
\label{table ablation scam}
\centering
\renewcommand{\arraystretch}{1.3}
\begin{tabular}{ccc|ccc}
\hline
\multirow{2}{*}{SCA} & \multirow{2}{*}{ReLU} & \multirow{2}{*}{GIM} & \multicolumn{3}{c}{RGB / Sonar}           \\ \cline{4-6} 
                     &                       &                      & SR          & PR          & NPR               \\ \hline 
                     &                       &                      & 69.3 / 40.9 & 73.5 / 54.6 & 75.9 / 51.6       \\
\checkmark           &                       &                      & 70.1 / 46.0 & 74.2 / 60.9 & 76.0 / 56.3       \\
\checkmark           & \checkmark            &                      & 70.3 / 48.1 & 74.6 / 64.8 & 76.6 / 59.6       \\
\checkmark           & \checkmark            & \checkmark           & 71.2 / 50.3 & 75.6 / 66.2 & 78.0 / 62.1       \\ \hline
\end{tabular}
\end{table}

%% file: tables/ablation_srst.tex
\begin{table}[]
\caption{Ablation of different components of SRST.}
\label{table ablation srst}
\centering
\renewcommand{\arraystretch}{1.3}
\begin{tabular}{ccc|ccc}
\hline
\multirow{2}{*}{SOT}    & \multirow{2}{*}{ToS}  & \multirow{2}{*}{WSD}      & \multicolumn{3}{c}{RGB / Sonar}               \\ \cline{4-6} 
                        &                       &                           & SR          & PR          & NPR               \\ \hline
\multicolumn{3}{c|}{RGB-baseline}                                           & 69.3 / 40.9 & 73.5 / 54.6 & 75.9 / 51.6       \\ \hline
\checkmark              &                       &                           & 70.3 / 41.4 & 73.8 / 55.5 & 76.2 / 51.8       \\
\checkmark              & \checkmark            &                           & 70.6 / 48.3 & 74.4 / 65.1 & 77.0 / 59.6       \\
\checkmark              & \checkmark            & \checkmark                & 71.2 / 50.3 & 75.6 / 66.2 & 78.0 / 62.1       \\ \hline
\end{tabular}
\end{table}

%% file: tables/ablation_layers.tex
\begin{table}[]
\caption{Ablation of different components of Insert layers.}
\label{table ablation layers}
\centering
\renewcommand{\arraystretch}{1.3}
\begin{tabular}{ccc|ccc}
\hline
\multicolumn{3}{c|}{layers} & \multicolumn{3}{c}{RGB / Sonar}     \\ \hline
4                    & 7                     & 10                   & SR            & PR            & NPR                 \\ \hline
                     &                       &                      & 69.3 / 40.9   & 73.5 / 54.6   & 75.9 / 51.6         \\
\checkmark           &                       &                      & 69.1 / 48.9   & 72.1 / 65.2   & 74.5 / 59.8         \\
\checkmark           & \checkmark            &                      & 70.4 / 49.4   & 74.0 / 65.7   & 77.1 / 60.2         \\
\checkmark           & \checkmark            & \checkmark           & 71.2 / 50.3   & 75.6 / 66.2   & 78.0 / 62.1         \\ \hline
\end{tabular}
\end{table}

%% file: figures/fig_rgbs_heatmap.tex
\begin{figure*}[]
	\centering
    \includegraphics[width=18cm]{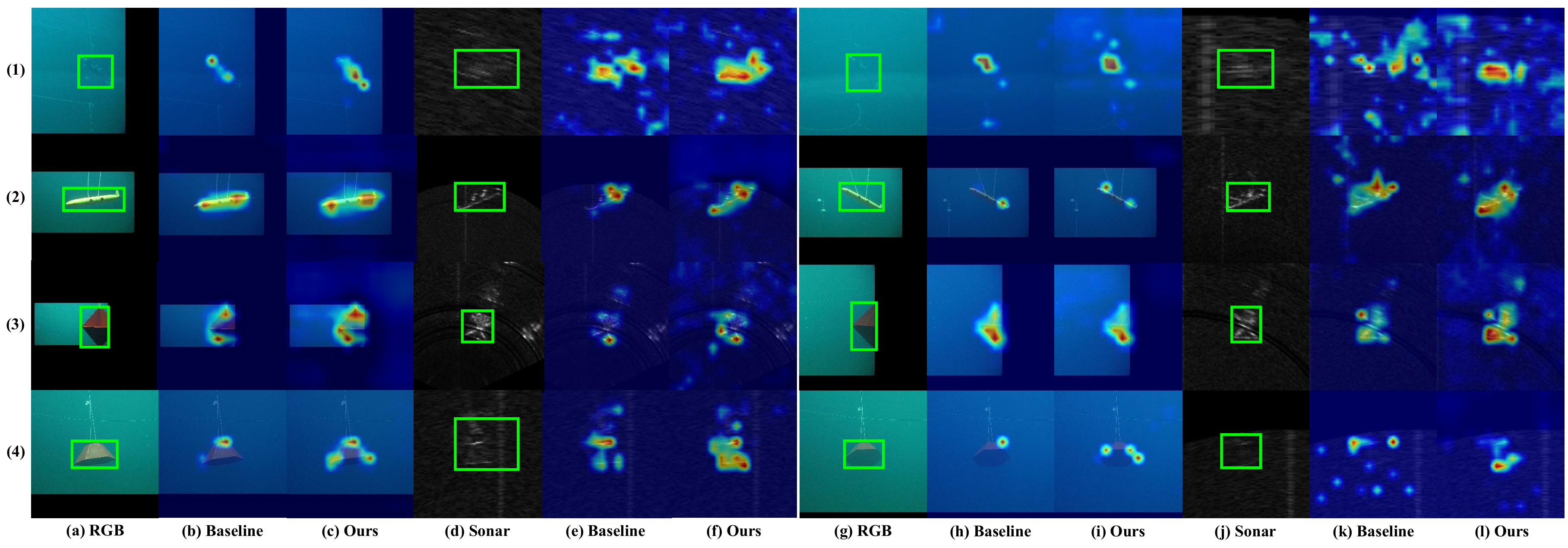}
	\caption{Visualization of heat maps of two modality backbone features produced by our method and the baseline approach. (a) (g) The RGB search area. (b) (h) The RGB heat maps of the baseline \cite{ostrack}. (c) (i) The RGB heat maps of our SCANet. (d) (j) The sonar search area. (e) (k) The sonar heat maps of the baseline \cite{ostrack}. (f) (l) The sonar heat maps of our SCANet. (1) fake\_person2 (\# 46 and 168). (2) UUV2 (\# 18 and 258). (3) octahedron1 (\# 22 and 188). (4) frustum1 (\# 47 and 95)}
	\label{fig: rgbs_heatmap}
\end{figure*}

%% file: figures/fig_rgb_trackingresults.tex
\begin{figure*}
	\centering
        \includegraphics[width=18cm]{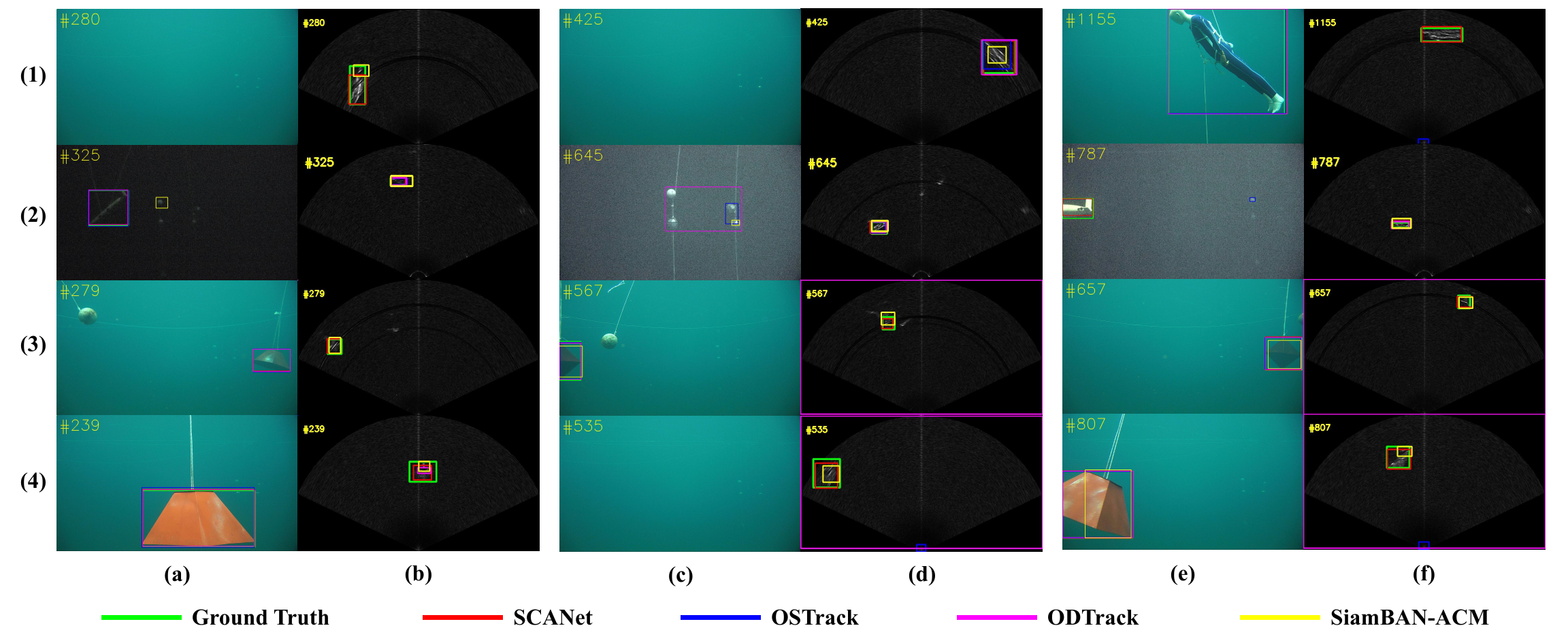}
	\caption{Visualized comparisons of SCANet with ODTrack \cite{odtrack}, OSTrack \cite{ostrack}, SiamBAN-ACM \cite{siambanacm} on four sequenses from RGBS50 dataset. (1) fake\_person4 sequence. (2) uuv9 sequence. (3) octahedron2
 sequence. (4) frustum6 sequence.}
	\label{fig: trackingresults}
\end{figure*}

%% file: sections/6_conclusion.tex
\section{Conclusion}

In this work, we introduce a novel multimodal tracking task, RGB-Sonar (RGB-S) tracking. It achieves accurate and stable tracking of underwater targets through the complementarity of underwater camera and sonar in terms of perceptual range and semantic richness. We first propose the first RGB-S tracking benchmark called RGBS50. The results of baseline methods show that the proposed RGBS50 is challenging to current SOT methods. Second, we propose an RGB-S tracker, SCANet, which includes two novel components: a spatial cross-attention module (SCAM) and a SOT data-based RGB-S simulation training method (SRST). The SCAM achieves the interaction between spatially misaligned RGB features and sonar features, and overcomes feature interference in misaligned spatial backgrounds. The SRST effectively trains the model using pseudo-data with similar semantic structures. Comprehensive experiments show that the proposed SCANet achieves state-of-the-art performance.

%% file: sections/acknowledgement.tex
\section{acknowledgement}
This research is funded by the National Natural Science Foundation of China, grant number 52371350, by the National Key Research and Development Program of China, grant number 2023YFC2809104, and by the National Key Laboratory Foundation of Autonomous Marine Vehicle Technology, grant number 2024-HYHXQ-WDZC03.